\documentclass[final]{static/l4dc2026}

\usepackage{physics}
\usepackage{enumitem}
\usepackage{amsmath,amssymb,amsfonts,pifont}
\usepackage{textcomp}
\usepackage{footnote}
\usepackage{array}    
\usepackage{tikz}
\usepackage{wrapfig} 
\makesavenoteenv{tabular}
\makesavenoteenv{table}
\usepackage{booktabs,bigstrut,rotating}
\usepackage{xcolor}
\usepackage[nolist, nohyperlinks, printonlyused, withpage, smaller]{acronym}
\usepackage{graphicx,import}    
\usepackage{subcaption}  
\usepackage{algorithm}
\usepackage{algorithmic}
\usepackage{enumitem}
\usepackage{tablefootnote}
\usepackage{caption}

\newcommand{\nablabx}{\nabla_{\!\bi{x}}}%
\newcommand{\cmark}{\textcolor[RGB]{0,120,0}{\scalebox{1.2}{\ding{51}}}}%
\newcommand{\xmark}{\textcolor[RGB]{120,0,0}{\scalebox{1.2}{\ding{55}}}}%

\newcommand{\rot}[1]{\begin{sideways}\shortstack{#1}\end{sideways}}

\newcommand{\smallsquare}[3]{%
	\tikz[baseline=-0ex]{
		\definecolor{customcolor}{rgb}{#1,#2,#3}%
		\draw[fill=customcolor,draw=none] (0,0) rectangle (1.4ex,1.4ex);
	}%
}

\newcommand{\smallsquareline}[3]{%
	\tikz[baseline=-0ex]{
		\definecolor{customcolor}{rgb}{#1,#2,#3}%
		\draw[fill=customcolor,draw=none] (0,0) rectangle (1.4ex,1.4ex);
		\draw[line width=1.5pt, color=customcolor] (-0.05,0.7ex) -- (1.8ex,0.7ex);
	}%
}

\newtheorem{assumption}{Assumption}

\title[Learning Dynamics from Input-Output Data with Hamiltonian GPs]{Learning Dynamics from Input-Output Data \\ with Hamiltonian Gaussian Processes}
\usepackage{times}

\author{%
	\Name{Jan-Hendrik Ewering}$^{1,2}$ \Email{jan-hendrik.ewering@imes.uni-hannover.de}\\
	\Name{Robin E. Herrmann}$^1$ \Email{robin.erik.herrmann@stud.uni-hannover.de}\\
	\Name{Niklas {Wahlstr{\"o}m}}$^2$ \Email{niklas.wahlstrom@it.uu.se}\\
	\Name{Thomas B. Sch{\"o}n}$^2$ \Email{thomas.schon@uu.se}\\
	\Name{Thomas Seel}$^1$ \Email{thomas.seel@imes.uni-hannover.de}%
	\AND
	\addr $^1$Leibniz Universit{\"a}t Hannover, 30823 Garbsen, Germany\\
	\addr $^2$Uppsala University, 751 05 Uppsala, Sweden
}

\begin{document}

\begin{acronym}
	\acro{rnn}[RNN]{Recurrent Neural Network}
	\acroplural{rnn}[RNN]{Recurrent Neural Networks}
	\acro{narx}[NARX-NN]{Nonlinear Autoregressive Exogenous Neural Network}
	\acroplural{narx}[NARX-NN]{Nonlinear Autoregressive Exogenous Neural Networks}
	\acro{gru}[GRU]{{Gated Recurrent Unit}}
	\acroplural{gru}[GRU]{{Gated Recurrent Units}}
	\acro{lstm}[LSTM]{Long Short-Term Memory}
	\acro{ann}[NN]{Neural Network}	
	\acroplural{ann}[NN]{Neural Networks}
	\acro{ffnn}[FFNN]{Feedforward Neural Network}
	\acroplural{ffnn}[FFNN]{Feedforward Neural Networks}
	\acro{pinn}[PINN]{Physics-Informed Neural Network}
	\acroplural{pinn}[PINN]{Physics-Informed Neural Networks}
	\acro{gp}[GP]{Gaussian Process}
	\acroplural{gp}[GPs]{Gaussian Processes}
	\acro{knn}[$K$NN]{$K$-Nearest-Neighbors}
	\acro{ilc}[ILC]{Iterative Learning Control}
	\acro{ili}[ILI]{Iterative Learning Identification}
	\acro{rls}[RLS]{Recursive Least Squares}
	\acro{rc}[RC]{Repetitive Control}
	\acro{rl}[RL]{{Reinforcement Learning}}
	\acro{daoc}[DAOC]{{Direct Adaptive Optimal Control}}
	\acro{ml}[ML]{maschinelles Lernen}
	\acro{lwpr}[LWPR]{{Locally Weighted Projection Regression}}
	\acro{svm}[SVM]{{Support Vector Machine}}
	\acro{mcmc}[MCMC]{{Markov Chain Monte Carlo}}
	\acro{ad}[AD]{{Automatic Differentiation}}
	\acro{gmm}[GMM]{{Gaussian Mixture Models}}
	\acro{rkhs}[RKHS]{{Reproducing Kernel Hilbert Spaces}}
	\acro{rbf}[RBF]{{Radial Basis Function}}
	\acro{rbfnn}[RBF-NN]{{Radial Basis Function Neural Network}}
	\acroplural{rbfnn}[RBF-NN]{{Radial Basis Function Neural Networks}}
	\acro{node}[{Neural} ODE]{{Neural Ordinary Differential Equation}}
	\acroplural{node}[{Neural} ODEs]{{Neural Ordinary Differential Equations}}
	\acro{pdf}[PDF]{Probability Density Function}
	\acro{pca}[PCA]{Principal Component Analysis}

	\acro{kf}[KF]{Kalman Filter}	
	\acro{ekf}[EKF]{Extended Kalman Filter}
	\acro{nekf}[NEKF]{Neural Extended Kalman Filter}
	\acro{ukf}[UKF]{Unscented Kalman Filter}
	\acro{pf}[PF]{Particle Filter}
	\acro{mhe}[MHE]{{Moving Horizon Estimation}}
	\acro{rpe}[RPE]{{Recursive Predictive Error}}
	\acro{slam}[SLAM]{{Simultaneous Location and Mapping}}
	
	\acro{mftm}[MFTM]{Magic Formula Tire Model}
	\acro{mbs}[MBS]{Multi-Body Simulation}
	\acro{lti}[LTI]{Linear Time-Invariant}
	\acro{cog}[COG]{Center Of Gravity}
	\acro{ltv}[LTV]{Linear Time-Variant}
	\acro{siso}[SISO]{Single Input Single Output}
	\acro{mimo}[MIMO]{Multiple Input Multiple Output}
	\acro{psd}[PSD]{Power Spectral Density}
	\acroplural{psd}[PSD]{Power Spectral Densities}
	\acro{cf}[CF]{Coordinate Frame}
	\acroplural{cf}[CF]{Coordinate Frames}
	\acro{phs}[PHS]{Port-Hamiltonian System}

	\acro{pso}[PSO]{Particle Swarm Optimization}
	\acro{sqp}[SQP]{Sequentielle Quadratische Programmierung}
	\acro{svd}[SVD]{Singular Value Decomposition}
	\acro{ode}[ODE]{Ordinary Differential Equation}
	\acroplural{ode}[ODE]{Ordinary Differential Equations}
	\acro{pde}[PDE]{Partial Differential Equation}
	
	\acro{nmse}[NMSE]{Normalized Mean Squared Error}
	\acro{mse}[MSE]{Mean Squared Error}
	\acro{rmse}[RMSE]{Root Mean Squared Error}
	\acro{wrmse}[wRMSE]{weighted \ac{rmse}}
	
	\acro{mpc}[MPC]{{Model Predictive Control}}
	\acro{nmpc}[NMPC]{Nonlinear Model Predictive Control}
	\acro{lmpc}[LMPC]{Learning Model Predictive Control}
	\acro{ltvmpc}[LTV-MPC]{\acl{ltv} Model Predictive Control}
	\acro{ndi}[NDI]{Nonlinear Dynamic Inversion}
	\acro{ac}[AC]{Adhesion Control}
	\acro{esc}[ESC]{Electronic Stability Control}
	\acro{ass}[ASS]{Active Suspension System}
	\acro{trc}[TRC]{Traction Control}
	\acro{abs}[ABS]{Anti-Lock Brake System}
	\acro{gcc}[GCC]{Global Chassis Control}
	\acro{ebs}[EBS]{Electronic Braking System}
	\acro{adas}[ADAS]{Advanced Driver Assistance Systems}
	\acro{cm}[CM]{{Condition Monitoring}}
	\acro{hil}[HiL]{{Hardware-in-the-Loop}}
	\acro{siso}[SISO]{Single Input Single Output}
	\acro{mimo}[MIMO]{Multiple Input Multiple Output}

	\acro{irw}[IRW]{Independently Rotating Wheels}
	\acro{dirw}[DIRW]{Driven \acl{irw}}
	\acro{imes}[imes]{{Institute of Mechatronic Systems}}
	\acro{db}[DB]{Deutsche Bahn}
	\acro{ice}[ICE]{Intercity-Express}
	
	\acro{fmi}[FMI]{{Functional Mock-up Interface}}
	\acro{fmu}[FMU]{{Functional Mock-up Unit}}
	\acro{doi}[DOI]{{Digital Object Identifier}}

	\acro{ra}[RA]{Research Area}
	\acroplural{ra}[RA]{Research Areas}
	\acro{wp}[WP]{Work Package}
	\acroplural{wp}[WP]{Work Packages}
	
	\acro{fb}[FB]{Forschungsbereich}
	\acroplural{fb}[FB]{Forschungsbereiche}
	\acro{ap}[AP]{Arbeitspaket}
	\acroplural{ap}[AP]{Arbeitspakete}
	\acro{abb}[Abb.]{Abbildung}
	\acro{luis}[LUIS]{Leibniz Universit"at IT Services}
	
	\acro{res}[RES]{Renewable Energy Sources}
	\acro{pkw}[PKW]{Personenkraftwagen}
	\acro{twipr}[TWIPR]{{Three-Wheeled Inverted Pendulum Robot}}
	\acro{tlr}[TLR]{Two-Link Robot}
	
	\acro{pmcmc}[PMCMC]{Particle Markov Chain Monte Carlo}
	\acro{mcmc}[MCMC]{Markov Chain Monte Carlo}
	\acro{rbpf}[RBPF]{Rao-Blackwellized Particle Filter}
	\acro{pf}[PF]{Particle Filter}
	\acro{ps}[PS]{Particle Smoother}
	\acro{smc}[SMC]{Sequential Monte Carlo}
	\acro{csmc}[cSMC]{conditional SMC}
	\acro{mh}[MH]{Metropolis Hastings}
	\acro{em}[EM]{Expectation Maximization}
	\acro{slam}[SLAM]{Simultaneous Location and Mapping}
	\acro{dof}[DOF]{Degree of Freedom}
	\acroplural{dof}[DOF]{Degrees of Freedom}
	\acro{pg}[PG]{Particle Gibbs}
	\acro{pgas}[PGAS]{Particle Gibbs with Ancestor Sampling}
	\acro{mpgas}[mPGAS]{marginalized Particle Gibbs with Ancestor Sampling}
	\acro{hmm}[HMM]{Hidden Markov Model}
	
	\acro{emps}[EMPS]{Electro-Mechanical Positioning System}
	
	\acro{ilc}[ILC]{Iterative Learning Control}
	\acro{ddilc}[DD-ILC]{Data-Driven Iterative Learning Control}
	\acro{dilc}[DILC]{Dual Iterative Learning Control}
	\acro{iml}[IML]{Iterative Model Learning}
	\acro{noilc}[NO-ILC]{Norm-Optimal Iterative Learning Control}
	\acro{gilc}[G-ILC]{Gradient Iterative Learning Control}

	\acro{bilbo}[BILBO]{Balancing Intelligent Learning roBOt}

\end{acronym}

\newcommand{\qvec}[1]{\mathbf{#1}}
\newcommand{\qmat}[1]{\mathbf{#1}}
\newcommand{\idx}[1]{_{\mathrm{#1}}}

\newcommand{\qRealNumbers}{\mathbb{R}}
\newcommand{\qPositiveRealNumbers}{\mathbb{R}_{\geq 0}}
\newcommand{\qNaturalNumbersZero}{\mathbb{N}_{\geq 0}}
\newcommand{\qNaturalNumbersPos}{\mathbb{N}_{>0}}
\newcommand{\qNaturalNumbers}{\mathbb{N}}
\newcommand{\foralljinN}{\forall j \in \qNaturalNumbersZero, \quad}
\newcommand{\foralljgeqN}{\forall j \in \qNaturalNumbersPos, \quad}
\newcommand{\qBLT}{\mathcal{T}^{\mathrm{BLT}}}
\newcommand{\qNormal}{\mathcal{N}}

\newcommand\invChi{\mathop{\mbox{Scale-inv-$\chi^2$}}}

\newcommand{\qa}{\qvec{a}}
\newcommand{\qb}{\qvec{b}}
\newcommand{\qd}{\qvec{d}}
\newcommand{\qe}{\qvec{e}}
\newcommand{\qf}{\qvec{f}}
\newcommand{\qg}{\qvec{g}}
\newcommand{\qh}{\qvec{h}}
\newcommand{\qi}{\qvec{i}}
\newcommand{\qj}{\qvec{j}}
\newcommand{\qk}{\qvec{k}}
\newcommand{\ql}{\qvec{l}}
\newcommand{\qm}{\qvec{m}}
\newcommand{\qn}{\qvec{n}}
\newcommand{\qo}{\qvec{o}}
\newcommand{\qp}{\qvec{p}}
\newcommand{\qr}{\qvec{r}}
\newcommand{\qs}{\qvec{s}}
\newcommand{\qt}{\qvec{t}}
\newcommand{\qu}{\qvec{u}}
\newcommand{\qv}{\qvec{v}}
\newcommand{\qw}{\qvec{w}}
\newcommand{\qx}{\qvec{x}}
\newcommand{\qy}{\qvec{y}}
\newcommand{\qz}{\qvec{z}}
\newcommand{\qem}{\qvec{e}^{\mathrm{m}}}

\newcommand{\qA}{\qvec{A}}
\newcommand{\qB}{\qvec{B}}
\newcommand{\qC}{\qvec{C}}
\newcommand{\qD}{\qvec{D}}
\newcommand{\qE}{\qvec{E}}
\newcommand{\qF}{\qvec{F}}
\newcommand{\qG}{\qvec{G}}
\newcommand{\qH}{\qvec{H}}
\newcommand{\qI}{\qvec{I}}
\newcommand{\qJ}{\qvec{J}}
\newcommand{\qK}{\qvec{K}}
\newcommand{\qL}{\qvec{L}}
\newcommand{\qM}{\qvec{M}}
\newcommand{\qN}{\qvec{N}}
\newcommand{\qO}{\qvec{O}}
\newcommand{\qP}{\qvec{P}}
\newcommand{\qQ}{\qvec{Q}}
\newcommand{\qR}{\qvec{R}}
\newcommand{\qS}{\qvec{S}}
\newcommand{\qT}{\qvec{T}}
\newcommand{\qU}{\qvec{U}}
\newcommand{\qV}{\qvec{V}}
\newcommand{\qW}{\qvec{W}}
\newcommand{\qX}{\qvec{X}}
\newcommand{\qY}{\qvec{Y}}
\newcommand{\qZ}{\qvec{Z}}

\newcommand{\quv}{\bar{\qu}}
\newcommand{\qyv}{\bar{\qy}}
\newcommand{\qxv}{\bar{\qx}}
\newcommand{\qZero}{\qvec{0}}
\newcommand{\qdu}{\boldsymbol{\Delta}\qu}
\newcommand{\qdy}{\boldsymbol{\Delta}\qy}
\newcommand{\qep}{\hat{\qe}}
\newcommand{\qyp}{\hat{\qy}}

\newcommand{\qILCDesign}{\boldsymbol{D}}
\newcommand{\qIMLDesign}{\boldsymbol{\hat{D}}}
\newcommand{\qLIML}{\hat{\qL}}
\newcommand{\qeR}{\qe_\mathrm{R}}


\newcommand{\qff}{\boldsymbol{f}}
\newcommand{\qpf}{\boldsymbol{p}}
\newcommand{\qmf}{\boldsymbol{m}}
\newcommand{\qXf}{\boldsymbol{A}}

\newcommand{\qLift}{\mathcal{L}}
\newcommand{\qLu}{\boldsymbol{\mathcal{L}\idx{u}}}
\newcommand{\qLm}{\boldsymbol{\mathcal{L}\idx{m}}}
\newcommand{\qUb}{\bar{\qU}}
\newcommand{\qyh}{\hat{\qy}}
\newcommand{\qeh}{\hat{\qe}}
\newcommand{\qLh}{\hat{\qL}}
\newcommand{\qDeh}{\hat{\boldsymbol{\mathcal{D}}}}
\newcommand{\qDe}{{\boldsymbol{\mathcal{D}}}}
\newcommand{\qMt}{\tilde{\qM}}
\newcommand{\qPt}{\tilde{\qP}}
\newcommand{\qLt}{\tilde{\qL}}
\newcommand{\qQt}{\tilde{\qQ}}
\newcommand{\qSt}{\tilde{\qS}}
\newcommand{\qWt}{\tilde{\qW}}
\newcommand{\qRt}{\tilde{\qR}}
\newcommand{\qut}{\tilde{\qu}}
\newcommand{\qyt}{\tilde{\qy}}
\newcommand{\qet}{\tilde{\qe}}
\newcommand{\qzt}{\tilde{\qz}}
\newcommand{\qrt}{\tilde{\qr}}
\newcommand{\qXt}{\tilde{\qX}}

\newcommand{\qTb}{\bar{\qT}}
\newcommand{\qVb}{\bar{\qV}}

\newcommand{\qQh}{\hat{\qQ}}
\newcommand{\qSh}{\hat{\qS}}
\newcommand{\qXh}{\hat{\qX}}
\newcommand{\qWh}{\hat{\qW}}

\newcommand{\Lfunc}{\mathscr{L}}       
\newcommand{\Kfunc}{\mathscr{K}}
\newcommand{\KLfunc}{\Kfunc\negthinspace\negthinspace\Lfunc}

\newcommand{\qTwoNorm}[1]{\left\| {#1} \right\|_{2}}
\newcommand{\qInfNorm}[1]{\left\| {#1} \right\|_\infty}
\newcommand{\qOneNorm}[1]{\left\| {#1} \right\|\idx{1}}
\newcommand{\qNorm}[1]{\left\| {#1} \right\|}
\newcommand{\qGivenNorm}{\qNorm{\boldsymbol{\cdot}}}

\newcommand{\qred}[1]{{\color{red}#1}}
\newcommand{\imesorange}{E77B29}
\newcommand{\imesgruen}{C8D317}
\newcommand{\imesblauHundert}{00509B}
\newcommand{\imesblauZwanzig}{CCDCEB}
\newcommand{\imesblauVierzig}{99B9D8}

\newcommand{\eg}{e.\,g.,\,}
\newcommand{\ie}{i.\,e.,\,}
\newcommand*{\R}{\mathbb{R}}

\newcommand*{\MNIW}{\mathcal{MNIW}}
\newcommand*{\IW}{\mathcal{IW}}
\newcommand*{\T}{\mathcal{T}}
\newcommand*{\N}{\mathcal{N}}

\newcommand{\del}{\partial}
\newcommand{\bi}[1]{\boldsymbol{#1}}
\newcommand{\ur}[1]{\mathrm{#1}}
\newcommand{\cali}[1]{\mathcal{#1}}

\newcommand{\ubar}[1]{\underaccent{\bar}{#1}}
\newcommand{\ToDo}[1]{\todo[size=\tiny]{#1}}

\newcommand{\spans}[1]{\spanop\left( #1 \right)}

\newcommand{\ToDos}[1]{\todo[size=\tiny]{#1}}


	\maketitle
	
	\begin{abstract}%
		Embedding non-restrictive prior knowledge, such as energy conservation laws, into learning methods is a key motive to construct physically consistent dynamics models from limited data, relevant for, \eg model-based control. 
		Recent work incorporates Hamiltonian dynamics into \acp{gp} to obtain uncertainty-quantifying, energy-consistent models, but these methods rely on---rarely available---velocity or momentum data.
        In this paper, we study dynamics learning using Hamiltonian \acp{gp} and focus on learning solely from input–output data, without relying on velocity or momentum measurements. 
        Adopting a non-conservative formulation, energy exchange with the environment, \eg through external forces or dissipation, can be captured. 
		We provide a fully Bayesian scheme for estimating probability densities of unknown hidden states, \ac{gp} hyperparameters, as well as structural hyperparameters, such as damping coefficients. 
		The proposed method is evaluated in a nonlinear simulation case study and compared to a state-of-the-art approach that relies on momentum measurements. 
	\end{abstract}
	
	\begin{keywords}%
		Nonlinear System Identification, Gaussian Processes, Physics-informed Learning%
	\end{keywords}
	
	\section{Introduction}
	Learning the dynamics of a system is a key approach for enabling high-performance model-based control and system insight, even with limited prior model knowledge. 
	To this end, recent physics-informed machine learning approaches provide well-generalizing and data-efficient learning-based models by incorporating non-restrictive prior knowledge \citep{Geist.2021}. 
	Notable examples include embedding energy conservation laws into neural networks \citep{Cranmer.2019} or \acp{gp} \citep{Beckers.2022} to learn models that yield physically interpretable predictions. 
	In this context, choosing a \ac{gp} is a good idea for (at least) two reasons. 
	First, recent work has shown that kernel-based methods, such as \acp{gp}, are particularly well-suited to represent dynamical systems \citep{Ziegler.2024,Scampicchio.2025}. 
	Second, \acp{gp} inherently provide an uncertainty quantification of the predictions that can benefit the application, \eg through stochastic control or safe learning \citep{Brunke.2022}. 
	
	In particular, fusing Hamiltonian inductive biases into \acp{gp} is a vital approach for obtaining uncertainty-quantifying dynamics models while ensuring physically plausible, energy-consistent predictions. 
	However, to the best of our knowledge, all existing work relies on measurements of the entire system state, meaning that, \eg beyond position measurements, momentum or velocity data is required for training, which is a restrictive assumption in many practical settings. 
    While ad hoc approaches approximate velocities from position data via numerical differentiation, this is highly sensitive to measurement noise \citep{Chartrand.2011}. 
    Moreover, generalized momenta, required within Hamiltonian mechanics, cannot generally be constructed from position data.

	In this article, we consider learning of non-conservative dynamics with Hamiltonian \ac{gp}s. 
    In contrast to previous research---which relies on measurements of momenta or velocities---we address the more realistic problem setting of learning from input-output data only. 
	We propose a novel Hamiltonian \ac{gp} model, which is linear in its parameters, features simple closed-form gradient expressions, and exhibits computational complexity independent of the training data dimension, thereby improving its applicability in practical settings. 
	For learning the model, we present a fully Bayesian system identification scheme that estimates probability densities of unknown latent states, \ac{gp} hyperparameters, as well as structural hyperparameters, such as damping coefficients.

	\section{Related Work}

	Integrating continuous-time dynamics into learning-based models has attracted significant attention over the last decade, with early work focusing on learning a vector-valued flow map that determines the evolution of a system's state, \eg using neural ordinary differential equations \citep{Chen.2018}. 
	A more contemporary approach is to compute this flow map from an approximation of the system's Hamiltonian or Lagrangian, which reflects the exchange of energy within a system. 
	In these approaches, a learning-based model, \eg a neural network \citep{Cranmer.2019,Lutter.2019b,Hansen.2025} or a \ac{gp} \citep{Evangelisti.2022b,Giacomuzzo.2024,Dai.2024,Beckers.2022} approximates the scalar Hamiltonian or Lagrangian. 
	By applying deterministic operations to the model, such as the Euler-Lagrange equations \citep{Evangelisti.2022b,Giacomuzzo.2024,Dai.2024} or a Hamiltonian system structure \citep{Beckers.2022,Greydanus.2019}, a physically consistent representation---obeying the underlying energy conservation laws by construction---is obtained.  
    Importantly, incorporating this ``algebraic'' physics information \citep{Watson.2025} contrasts with simulation-based learning methods, which feature a physics loss based on a known partial differential equation \citep{Raissi.2019}. 

    \begin{table}[h]
		\centering
		\caption{\centering Comparison of recent Hamiltonian \acp{gp} ($\bi{q}$: positions/coordinates; $\bi{p}$: momenta)}
		\selectfont\footnotesize
		\begin{tabular}{l|lllllll}
			\toprule
			& \rot{Exogenous \\ inputs} & \rot{Dissipative \\ systems} & \rot{Variables required \\ for training} & \rot{Reduced-rank \ac{gp}  \\ (comp. efficiency)} & \rot{\ac{gp} \\ hyperparameter \\ learning} & \rot{Structural \\ hyperparameter \\ learning \\ (\eg damping)} \\
			\midrule
			\cite{Bertalan.2019}   & \xmark & \xmark & $\dot{\bi{q}},\dot{\bi{p}}$  & \xmark & \xmark & \xmark \\
			\cite{Rath.2021,Offen.2022}  & \xmark & \xmark & $\bi{q},{\bi{p}}$  & \xmark & frequentist & \xmark \\
			\cite{Tanaka.2022}   & \xmark & \cmark & $\bi{q},{\bi{p}}$  & \cmark & frequentist & frequentist \\
			\cite{Ensinger.2023}   & \xmark & \xmark & $\bi{q},{\bi{p}}$  & \xmark & frequentist & \xmark \\
			\cite{Beckers.2022}   & \cmark & \cmark & $\bi{q},{\bi{p}}$  & \xmark & frequentist & frequentist \\
			\cite{Ross.2023}   & \xmark & \xmark & $\bi{q},{\bi{p}}$  & \cmark & frequentist & \xmark \\
			\cite{Hu.2025}   & \xmark & \xmark & $\dot{\bi{q}},\dot{\bi{p}}$  & \xmark & frequentist & \xmark \\
			Proposed   & \cmark & \cmark & $\bi{q}$  & \cmark & Bayesian & Bayesian\\
			\bottomrule
		\end{tabular}%
		\label{tab:lit_rev}%
	\end{table}%
    
	Considering Hamiltonian-based \ac{gp} approaches (see Table\,\ref{tab:lit_rev}), existing work often models the underlying system as conservative \citep{Bertalan.2019,Rath.2021}, meaning that there is no energy exchange with the environment.
    More recent methods consider energy dissipation \citep{Tanaka.2022} and exogenous inputs, such as driving forces and torques \citep{Beckers.2022,Li.2024b}. 
	Although this represents a step toward realistic problem settings, all existing work on Hamiltonian \acp{gp} relies on velocity or momentum measurements, which is a restrictive assumption in many applications. 
	This issue has been noted by \cite{Hansen.2025}, who provide a \textit{deterministic} neural network-based approach for energy-consistent learning from position data. 
    While uncertainty quantification can be achieved through various frameworks---such as Bayesian neural networks \citep{Jospin.2022} or deep ensembles \citep{Lakshminarayanan.2017}---these probabilistic approaches do not inherently yield physically consistent dynamics. 
	Hence, an open problem is to learn energy-consistent and non-conservative system models from input-output data while also providing an uncertainty quantification.

	Moreover, all research on Hamiltonian \acp{gp} that considers hyperparameter learning relies on a frequentist assumption and does not provide uncertainty quantification over structural model properties, such as \ac{gp} hyperparameters or damping coefficients. 
	Lastly, predicting with \acp{gp} can be computationally costly when the number of training data points exceeds the ``small-data'' regime. 
	In this light, only a few studies \citep{Tanaka.2022,Ross.2023} address the computational burden with approximations to make the practical application of Hamiltonian \acp{gp} feasible.

	\section{Problem Formulation}\label{sec:problem}
	We are concerned with learning physically consistent state-space models from input-output data.  
	To this end, we consider systems whose dynamics can be described by forced Hamiltonian mechanics. 
	Building on classical mechanics, this formulation provides a modeling paradigm that allows for describing energy storage and dissipation within a system, as well as energy exchange across systems, in a consistent and interpretable fashion. 
	Formally, consider a continuous-time state-space system
	\begin{equation}\label{eq:problem}
            \dot{\bi{x}} = \left(\bi{J}(\bi{x}) - \bi{R}(\bi{x})\right) \nablabx  H(\bi{x}) + \bi{G}(\bi{x}) \bi{u} + \bi{w}\, ,  \qquad \qquad  \bi{y} = \bi{g}(\bi{x}) + \bi{e} \, , 
	\end{equation}
	where the energy is described by the \textit{unknown} Hamiltonian function $H: \Omega \rightarrow \mathbb{R}$. The \textit{unknown} system state $\bi{x} = [\bi{q}^\top, \bi{p}^\top]^\top \in \Omega \subset \mathbb{R}^{n_x} $---consisting of the generalized coordinates $\bi{q} \in \mathbb{R}^{n_q}$ and the conjugate momenta of the system $\bi{p} \in \mathbb{R}^{n_p}$---is driven by the exogenous inputs $\bi{u} \in \mathbb{R}^{n_u}$, and observed through the outputs $\bi{y} \in \mathbb{R}^{n_y}$. 
	We consider a stochastic setting with normally distributed process noise $\bi{w} \sim  \mathcal{N}(\bi{w} \mid \bi{0},\widetilde{\bi{\Sigma}}_{w})$ and measurement noise $\bi{e} \sim   \mathcal{N}(\bi{e} \mid \bi{0}, \widetilde{\bi{\Sigma}}_e)$, respectively.

    \begin{assumption}
		The covariances $\widetilde{\bi{\Sigma}}_w$, $\widetilde{\bi{\Sigma}}_e$, and the map $\bi{g} : \mathbb{R}^{n_x} \rightarrow \mathbb{R}^{n_y}$ are assumed to be known.
	\end{assumption}

    While other noise models could be used, it is often sufficient in practical settings to consider Gaussian noise. 
	The energy exchange in the system is described by the skew-symmetric interconnection matrix $\bi{J}\in \mathbb{R}^{n_x \times n_x}$, the symmetric, positive semi-definite dissipation matrix $\bi{R}\in \mathbb{R}^{n_x \times n_x}$, $\bi{R} = \bi{R}^\top \succeq 0$, and the input matrix $\bi{G}\in \mathbb{R}^{n_x \times n_u}$. 
    
    \begin{assumption}
		The parametric structures of the matrices $\bi{J}(\bi{x})$, $\bi{R}(\bi{x})$, and $\bi{G}(\bi{x})$, \ie the potentially state-dependent patterns of entries, are assumed to be known, but the parameters themselves are \textit{unknown}. 
	\end{assumption}

    While the \textit{unknown} Hamiltonian captures the potentially highly nonlinear energy landscape of a given system, the matrices $\bi{J}$, $\bi{R}$, and $\bi{G}$ typically encode the basic physical system topology, such as kinematic relationships, which are often available from coarse system understanding. 
    
	Specifically, the problem considered in this work is to learn an uncertainty-quantifying model, obeying the underlying physics \eqref{eq:problem}, from sampled input-output data $\{\bi{u}_t,\bi{y}_t\}_{t=0}^{T}$. 
	This amounts to estimating the joint conditional distribution\footnote{To improve readability, we sometimes refer to a probability density function as a distribution, and use the short-hand notation $\bi{x}_{0:T}:= \{ \bi{x}_t \}_{t=0}^T$ to describe time-series data.} $p(\bi{x}_{0:T}, \bi{\xi} | \bi{u}_{0:T}, \bi{y}_{0:T})$ of hidden states and model parameters~$\bi{\xi}$---describing the Hamiltonian $H$, as well as the matrices $\bi{J}(\bi{x})$, $\bi{R}(\bi{x})$, and $\bi{G}(\bi{x})$---in a fully Bayesian setting.

	\section{Reduced-Rank Hamiltonian Gaussian Processes}\label{sec:model}
	To construct a physically consistent, uncertainty-quantifying system representation, we model the Hamiltonian function as a zero-mean \acl{gp} (\acs{gp}), \ie ${H}(\bi{x}) \sim  \mathcal{GP}(0, \kappa(\bi{x}, \bi{x}'))$, with covariance (kernel) function $k(\bi{x}, \bi{x}')$ \citep{Rasmussen.2005}, as proposed by \cite{Beckers.2022}. 
	Importantly, not the Hamiltonian function itself, but its gradient is needed for prediction within the model structure \eqref{eq:problem}. 
	Therefore, exploiting that \acp{gp} are closed under linear operations, we model the gradient of the Hamiltonian as
	\begin{equation}\label{eq:gradient_Hamiltonian}
		\nablabx   {H}(\bi{x}) \sim \mathcal{GP}(\bi{0}, \nablabx   \nabla_{\bi{x}'} \kappa(\bi{x}, \bi{x}')) \, ,
	\end{equation}
    where applying the differential operators to a differentiable scalar kernel yields a valid positive-definite matrix-valued kernel \citep{Beckers.2022}. 
	However, a common drawback of \acp{gp} is that their computational complexity and memory requirements scale poorly with the number of training data points, which becomes even more severe in \eqref{eq:gradient_Hamiltonian} because the partial derivative induces a multi-output \ac{gp}. 
	Instead, for efficient learning in practical settings, it is desirable that a model features \textit{(i)} a computational complexity independent of the training data dimension, \textit{(ii)} a linear representation to facilitate closed-form learning, and \textit{(iii)}~access to computationally simple gradient expressions. 
	
	Interestingly, the reduced-rank \ac{gp} presented by \cite{Solin.2020} exhibits all of these properties, which is why it has been exploited in various related papers on parameter learning and state inference \citep{Svensson.2017,Ewering.2026}. 
    To retain conciseness, we refer to \cite{Svensson.2017} for a justification of the approach and introduce only the main concept subsequently. 
	Loosely speaking, in the reduced-rank \ac{gp} \citep{Solin.2020}, a covariance function is approximated with a finite-dimensional eigenfunction expansion by encoding the kernel's spectral density $S$ in the frequency domain. 
	The chosen kernel is described by
	\begin{equation}\label{eq:kernel_approx}
		\kappa(\bi{x}, \bi{x}') \approx \sum_{k=1}^{M} S(\sqrt{\varrho_k}) \phi_k(\bi{x}) \phi_k(\bi{x}')\, ,
	\end{equation}
	where $\phi_k : \Omega \rightarrow \mathbb{R}$ are eigenfunctions of the Laplace operator on the domain $\Omega = [-L_1, L_1] \times \dots \times [-L_{n_x}, L_{n_x}]$ where the system state resides, and $\varrho_k$ are the corresponding eigenvalues. 
	For this rectangular domain, the eigenfunctions have a closed form, that is
	\begin{equation}\label{eq:basis_functions}
		\phi_{k}(\bi{x}) = \prod_{i=1}^{n_x} \frac{1}{\sqrt{L_i}} \sin \left(\frac{\pi j_{k,i}\left(x_{i}+L_i\right)}{2 L_i}\right)\, , \qquad \varrho_{k} = \sum_{i=1}^{n_x}\left(\frac{\pi j_{k,i}}{2 L_i}\right)^2\, ,
	\end{equation}
	where $\bi{x} = [x_1,\hdots,x_{n_x}]^\top$, and the indices $j_{k,i}$ determine the frequency of the corresponding eigenfunction \citep{RiutortMayol.2023}. 
	Predictions of ${H}(\bi{x}) \sim \mathcal{GP}(0, \kappa(\bi{x}, \bi{x}'))$ can be performed with the reduced-rank \ac{gp} at computational complexity $\mathcal{O}(M)$ using the basis function expansion
	\begin{equation}\label{eq:rr_gp_pred}
		\widehat{H}(\bi{x}) = \sum_{k=1}^{M} a_k \phi_k(\bi{x}) = \bi{a}^\top  \bi{\phi}(\bi{x}) \, ,
	\end{equation}
	where the vector-valued function $\bi{\phi}(\bi{x}) = [\phi_1(\bi{x}), \hdots, \phi_M(\bi{x})]^\top$, and the weights $\bi{a} = [a_1, \hdots, a_M]^\top$ follow a distribution $\mathcal{N} ( \bi{0}, \bi{V})$, with $\bi{V} = \mathrm{diag}(S(\sqrt{\varrho_1}),\hdots,S(\sqrt{\varrho_M}))$. 
	Please note that we use the superscript $\widehat{\square}$ to denote the learned counterparts of the quantities in \eqref{eq:problem}. 
	The reduced-rank \ac{gp} converges to the exact \ac{gp} in the limit $M, L_1, \hdots, L_{n_x} \rightarrow \infty$ \citep{Solin.2020}. 
	Given this, we model the Hamiltonian gradient $\nablabx {H}$ by taking the partial derivative of \eqref{eq:rr_gp_pred}, that is
	\begin{equation}\label{eq:probab_model}
		\begin{aligned}
			\nablabx   \widehat{H}(\bi{x}) = \nablabx   \left( \bi{a}^\top \bi{\phi}(\bi{x}) \right) = \bi{D}_{\bi{\phi}}(\bi{x}) \bi{a}  \, ,
		\end{aligned}
	\end{equation}
	where $\bi{D}_{\bi{\phi}}(\bi{x}) \in \mathbb{R}^{n_x \times M}$ is the closed-form Jacobian of the transposed basis function vector $\bi{\phi}^\top(\bi{x})$. 
	Note that this model is still linear in its parameters $\bi{a}$, which is convenient for computationally efficient training and prediction.  
	Unfortunately, we do not have direct access to Hamiltonian measurements for learning. 
    Instead, we use particle Gibbs \citep{Svensson.2017} to decouple learning the Hamiltonian from inferring unknown hidden variables (see Section\,\ref{sec:inference_learning}). 
    For this, we model the current Hamiltonian gradient $\nablabx \widehat{H}(\bi{x})$ at time $t$ as an auxiliary quantity $\bi{h}_{t} \in \mathbb{R}^{n_x}$, \ie
	\begin{equation}\label{eq:probab_model2}
		\begin{aligned}
			\bi{h}_t  =  \left. \nablabx   \widehat{H}(\bi{x}) \right|_{\bi{x} = \bi{x}_t} + \bi{v}_t \, , \qquad \bi{v}_t \sim \mathcal{N}(\bi{0}, \sigma^2 \mathrm{\mathbf{I}}) \, ,
		\end{aligned}
	\end{equation}
    which is estimated via \ac{smc}. 
	For conjugacy reasons and to describe the model \eqref{eq:probab_model} \& \eqref{eq:probab_model2} with a single distribution \citep{Svensson.2017,Berntorp.2021}, we express the \ac{gp} prior as a zero-mean multivariate normal $\mathcal{N}(\bi{a} | \bi{0},\sigma^2 \bi{V})$, where the scale $\sigma^2 $ reflects the noise, and the diagonal covariance $\bi{V}$ encodes the spectral density of the chosen kernel. 
	While any isotropic kernel, \ie $\kappa(\bi{x}, \bi{x}') = \kappa( || \bi{x} -  \bi{x}' ||)$,  can be employed, we resort to using a common squared exponential kernel
	\begin{equation}
		\kappa_{\mathrm{se}}(\bi{x}, \bi{x}') = \sigma_f^2 \exp\left( -\frac{\lVert \bi{x} - \bi{x'} \rVert^2}{2\ell^2} \right) \, , \qquad S_{\mathrm{se}}(\omega) = \sigma_f^2 (2 \pi \ell^2)^{n_x/2} \exp \left(-\frac{ \ell^2 \omega^2}{2}\right) \, ,
	\end{equation}
	in the following, with Euclidean norm $\norm{\cdot}$, spectral density $S_{\mathrm{se}}(\cdot)$, as well as $\sigma_f^2$ and length scale $\ell$ as hyperparameters $\bi{\vartheta}_{\mathrm{K}} := \{\sigma_f^2, \ell\}$. 
    Please note that using isotropic kernels is a valid assumption, provided we expect the Hamiltonian's smoothness not to change drastically from one region of state space to another. 
	
	The noise parameter $\sigma^2$ is unknown and must be estimated along with the basis function coefficients $\bi{a}$. 
	To this end, we set an inverse Gamma prior $\mathcal{IG}(\sigma^2| {\psi}, \nu)$ on $\sigma^2$ \citep{Murphy.2007},
	with scale $\psi$ and degrees of freedom $\nu$, such that the overall \ac{gp} prior for parameters $\bi{\theta} = \{\bi{a}, \sigma^2\}$ becomes 
	\begin{equation} \label{eq:MdlStr_MNIW_Prior}
		\begin{aligned}
			\bi{a}, \sigma^2 \sim & \, \mathcal{NIG}(\bi{a}, \sigma^2| \bi{0}, \bi{V}, {\psi}, \nu)  =\mathcal{N}(\bi{a}|\bi{0}, \sigma^2 \bi{V}  )\mathcal{IG}(\sigma^2| {\psi}, \nu) \, .
		\end{aligned}
	\end{equation}
	For further details on the distributions employed in this reduced-rank \ac{gp} model, we refer to the Supplementary Material\,\ref{sup:restricted_exp_family} and \citet{Volkmann.2025}. 
	Using the gradient approximation $\nablabx   \widehat{H}(\bi{x})$,
	the state transition function of a non-conservative Hamiltonian \ac{gp} model can now be constructed as
	\begin{equation}\label{eq:contin_model}
		\dot{\bi{x}}  = \left(\widehat{\bi{J}}(\bi{x}) - \widehat{\bi{R}}(\bi{x})\right) \nablabx   \widehat{H}(\bi{x}) + \widehat{\bi{G}}(\bi{x})\bi{u} + \bi{w} \, ,
	\end{equation}
	where energy is transferred across the system boundary by the exogenous inputs $\bi{u}$, and dissipation is reflected through the matrix $\widehat{\bi{R}}(\bi{x})$. 
	To align with sampled data, a suitable integration scheme can be applied for discretizing the continuous-time dynamics \eqref{eq:contin_model}. 
	
	To learn the proposed model, the (hyper-)parameters $\bi{\xi} := \{\bi{\theta}, \bi{\vartheta}_{\mathrm{K}}, \bi{\vartheta}_{\mathrm{S}}\}$ must be determined. 
	In particular, we have \textit{(i)} the Hamiltonian model parameters $\bi{\theta}$, \textit{(ii)} the \ac{gp} kernel hyperparameters $\bi{\vartheta}_{\mathrm{K}}$, 
	and \textit{(iii)} the structural hyperparameters $\bi{\vartheta}_{\mathrm{S}}$, \ie the unknown entries of $\widehat{\bi{J}}(\bi{x})$, $\widehat{\bi{R}}(\bi{x})$, and $\widehat{\bi{G}}(\bi{x})$.
	
	\section{Bayesian Inference and Learning from Input-Output Data}\label{sec:inference_learning}
	
	This section introduces a Bayesian inference and learning scheme for estimating---under the given model structure \eqref{eq:contin_model}---the joint distribution $p(\bi{z}_{0:T},  \bi{\xi} | \bi{u}_{0:T}, \bi{y}_{0:T})$ 
	of latent variables $\bi{z}_{t} = \{\bi{x}_{t}, \bi{h}_t\}$ and (hyper-)parameters $\bi{\xi}$ from input-output\footnote{As commonly done in Bayesian estimation literature \citep{Sarkka.2023}, the dependence on exogenous inputs $\bi{u}_{0:T}$ is not explicitly stated for the remainder of the paper to avoid notational clutter.}  data $\{\bi{u}_t,\bi{y}_t\}_{t=0}^{T}$. 
	For a fully Bayesian treatment, we model the parameters $\bi{\theta}$ and the hyperparameters $\bi{\vartheta} := \{\bi{\vartheta}_{\mathrm{K}}, \bi{\vartheta}_{\mathrm{S}}\}$ as random variables and set a prior $p(\bi{\theta} | \bi{\vartheta}_{\mathrm{K}})$ and a hyper-prior $p(\bi{\vartheta})$, respectively. 
	
	Given this potentially high-dimensional target distribution, performing state inference and parameter learning can be challenging. 
	To tackle this task, we rely on a \ac{pmcmc} approach \citep{Andrieu.2010} and break down the overall problem into simpler sub-problems using a particle Gibbs scheme \citep{Lindsten.2014,Volkmann.2025}. The overall inference and learning method is illustrated in Figure\,\ref{fig:overview_chart} and detailed in Algorithm\,\ref{alg:Bayesian_inference_learning}. 
	Loosely speaking, the learning procedure consists of sequentially sampling from the conditional distributions
	\begin{enumerate}[label=\roman*., itemsep=0.1pt]
		\item of latent variables $\bi{z}_{0:T}$, given the measurements $\bi{y}_{0:T}$ and (hyper-)parameters $\bi{\xi}$, and \label{step_1}
		\item of (hyper-)parameters $\bi{\xi} = \{ \bi{\theta}, \bi{\vartheta}_{\mathrm{K}}, \bi{\vartheta}_{\mathrm{S}} \}$, given the latent variables $\bi{z}_{0:T}$. \label{step_2}
	\end{enumerate}
	
	To perform Step\,\ref{step_1}, we employ a conditional \ac{smc} procedure \citep{Lindsten.2014}. 
	It provides samples from the density $p(\bi{z}_{0:T} | \bi{y}_{0:T}, \bi{\xi})$, and resembles a standard particle filter with one trajectory fixed to a previous reference trajectory. 
    In Step\,\ref{step_2}, finding the posterior density of $\bi{\theta}$ is done in closed form, exploiting the conjugate prior and the parameter-linearity of the model. 
	The posterior density of $\bi{\vartheta} = \{\bi{\vartheta}_{\mathrm{K}}, \bi{\vartheta}_{\mathrm{S}}\}$ is targeted by \ac{mh} steps within the particle Gibbs scheme. 

	\begin{remark}
		The proposed inference and learning method resembles the particle Gibbs scheme for \ac{gp} learning by \cite{Svensson.2017}. 
		Algorithm\,\ref{alg:Bayesian_inference_learning} thus represents a valid \ac{pmcmc} sampler that is guaranteed to asymptotically sample from the true distribution $p(\bi{z}_{0:T},  \bi{\xi} | \bi{y}_{0:T})$, as the number of iterations $k\rightarrow\infty$ \citep{Andrieu.2010}.
	\end{remark}
	
	\begin{remark}
		While we focus on \textit{Bayesian} inference and learning in this paper, the reduced-rank Hamiltonian \ac{gp} model \eqref{eq:contin_model} also enables efficient learning in a \textit{frequentist} setting, exploiting the closed-form expressions for gradient-based hyperparameter optimization. 
		The computational complexity of learning the covariance function parameters is $\mathcal{O}(TM^2)$ for initialization and $\mathcal{O}(M^3)$ per evaluation of the marginal likelihood and its gradient \citep{Solin.2020}.
		This compares to a complexity of $\mathcal{O}(T^3)$ for each optimizer step using an exact \ac{gp} \citep{Rasmussen.2005}.
	\end{remark}

    \vspace{-5mm}
	\noindent
	\begin{minipage}[t]{0.40\textwidth}
		\centering
		\vspace{12pt} 
		{\fontsize{7pt}{7pt}\selectfont
			\resizebox{1\textwidth}{!}{
\begingroup%
  \makeatletter%
  \providecommand\color[2][]{%
    \errmessage{(Inkscape) Color is used for the text in Inkscape, but the package 'color.sty' is not loaded}%
    \renewcommand\color[2][]{}%
  }%
  \providecommand\transparent[1]{%
    \errmessage{(Inkscape) Transparency is used (non-zero) for the text in Inkscape, but the package 'transparent.sty' is not loaded}%
    \renewcommand\transparent[1]{}%
  }%
  \providecommand\rotatebox[2]{#2}%
  \newcommand*\fsize{\dimexpr\f@size pt\relax}%
  \newcommand*\lineheight[1]{\fontsize{\fsize}{#1\fsize}\selectfont}%
  \ifx\svgwidth\undefined%
    \setlength{\unitlength}{177bp}%
    \ifx\svgscale\undefined%
      \relax%
    \else%
      \setlength{\unitlength}{\unitlength * \real{\svgscale}}%
    \fi%
  \else%
    \setlength{\unitlength}{\svgwidth}%
  \fi%
  \global\let\svgwidth\undefined%
  \global\let\svgscale\undefined%
  \makeatother%
  \begin{picture}(1,0.80225989)%
    \lineheight{1}%
    \setlength\tabcolsep{0pt}%
    \put(0,0){\includegraphics[width=\unitlength,page=1]{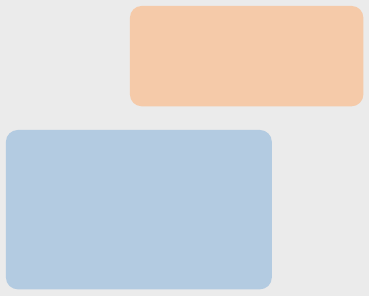}}%
    \put(0.36587205,0.74300336){\color[rgb]{0,0,0}\makebox(0,0)[lt]{\lineheight{1.10000002}\smash{\begin{tabular}[t]{l}\textbf{State inference}\end{tabular}}}}%
    \put(0.37916678,0.468771){\color[rgb]{0,0,0}\makebox(0,0)[lt]{\lineheight{1.10000002}\smash{\begin{tabular}[t]{l}$K$ iterations\end{tabular}}}}%
    \put(0.03053659,0.40582718){\color[rgb]{0,0,0}\makebox(0,0)[lt]{\lineheight{1.10000002}\smash{\begin{tabular}[t]{l}\textbf{Parameter learning}\end{tabular}}}}%
    \put(0.01503011,0.73056773){\color[rgb]{0,0,0}\makebox(0,0)[lt]{\lineheight{1.10000002}\smash{\begin{tabular}[t]{l}Input-output data\\$\{\bi{u}_t,\bi{y}_t\}_{t=0}^{T}$\end{tabular}}}}%
    \put(0.97379061,0.53925878){\color[rgb]{0,0,0}\makebox(0,0)[rt]{\lineheight{1.10000002}\smash{\begin{tabular}[t]{r}\textit{Conditional \ac{smc}}\end{tabular}}}}%
    \put(0.9693714,0.43485033){\color[rgb]{0,0,0}\makebox(0,0)[rt]{\lineheight{1.10000002}\smash{\begin{tabular}[t]{r}Latent\\variable\\samples\\$\bi{z}_{0:T}[k]$ \end{tabular}}}}%
    \put(0.7253437,0.2257218){\color[rgb]{0,0,0}\makebox(0,0)[rt]{\lineheight{1.10000002}\smash{\begin{tabular}[t]{r}\textit{Metropolis Hastings, Sec.\,\ref{sec:methods_hyperparameters}}\end{tabular}}}}%
    \put(0.73013275,0.04009524){\color[rgb]{0,0,0}\makebox(0,0)[rt]{\lineheight{1.10000002}\smash{\begin{tabular}[t]{r}\textit{Closed form, Sec.\,\ref{sec:methods_parameters}}\end{tabular}}}}%
    \put(0.98445213,0.01752749){\color[rgb]{0,0,0}\makebox(0,0)[rt]{\lineheight{1.10000002}\smash{\begin{tabular}[t]{r}\textit{Particle Gibbs}\end{tabular}}}}%
    \put(0,0){\includegraphics[width=\unitlength,page=2]{overview_chart.pdf}}%
    \put(0.58110968,0.69772156){\color[rgb]{0,0,0}\makebox(0,0)[lt]{\lineheight{1.10000002}\smash{\begin{tabular}[t]{l}$\bi{z}_{0:T}[k] \sim$ \\$p(\bi{z}_{0:T} | \bi{y}_{0:T}, \bi{\theta}, \bi{\vartheta}_{\mathrm{S}})$\end{tabular}}}}%
    \put(0.03957026,0.33782643){\color[rgb]{0,0,0}\makebox(0,0)[lt]{\lineheight{1.10000002}\smash{\begin{tabular}[t]{l}$\bi{\vartheta}_{\mathrm{S}}[k] \sim \mathcal{MH} (\bi{\vartheta}_{\mathrm{S}} | \bi{z}_{0:T} )$\\$\bi{\vartheta}_{\mathrm{K}}[k] \sim \mathcal{MH} (\bi{\vartheta}_{\mathrm{K}} | \bi{z}_{0:T} )$\end{tabular}}}}%
    \put(0.03956701,0.14929017){\color[rgb]{0,0,0}\makebox(0,0)[lt]{\lineheight{1.10000002}\smash{\begin{tabular}[t]{l}$\bi{\theta}[k] \sim p(\bi{\theta}|\bi{z}_{0:T}, \bi{\vartheta}_{\mathrm{K}})$\\$\quad \qquad = \mathcal{NIG}(\bi{\theta} | \bi{m}^{+}, \bi{V}^{+}, \psi^{+}, \nu^{+})$\end{tabular}}}}%
    \put(0.02648346,0.56540137){\color[rgb]{0,0,0}\makebox(0,0)[lt]{\lineheight{1.10000002}\smash{\begin{tabular}[t]{l}Parameter \\samples\\$\bi{\theta}[k], \bi{\vartheta}_{\mathrm{S}}[k]$ \end{tabular}}}}%
    \put(0,0){\includegraphics[width=\unitlength,page=3]{overview_chart.pdf}}%
  \end{picture}%
\endgroup%
}}
		
		\captionof{figure}{Algorithm overview.}
		\label{fig:overview_chart}
	\end{minipage}%
	\hfill
	\begin{minipage}[t]{0.58\textwidth}
		\vspace{0pt} 
		\begin{algorithm}[H] 
			\SetAlgoVlined 
			\caption{Inference and Learning of Hamiltonian GP}
			\label{alg:Bayesian_inference_learning}
			\KwIn{Data $\{\bi{u}_t,\bi{y}_t\}_{t=0}^{T}$, state density $p(\bi{x}_0)$, \\ parameter prior $p(\bi{\theta}|\bi{\vartheta}_{\mathrm{K}})$, hyper-prior $p(\bi{\vartheta})$}
			\KwOut{$K$ samples of invariant distr. $p(\bi{z}_{0:T},  \bi{\xi} | \bi{y}_{0:T})$}
			
			Initialize $ \bi{z}_{0:T}[0] $ arbitrarily ;
			
			Draw $ \bi{\vartheta}[0]\: | \: \bi{z}_{0:T}[0]$ \tcp*{Sec.\,\ref{sec:methods_hyperparameters}}
			
			Draw $\bi{\theta}[0] \: | \: \bi{z}_{0:T}[0], \bi{\vartheta}[0]$ \tcp*{Sec.\,\ref{sec:methods_parameters}}
			
			\For{$k = 1$ \KwTo $K$}{
				Draw $\bi{z}_{0:T}[k] \: | \: \bi{y}_{0:T}, \bi{z}_{0:T}[k-1], \bi{\theta}[k-1], \bi{\vartheta}_{\mathrm{S}}[k-1]$ 
				
				Draw $\bi{\vartheta}[k] \: | \: \bi{z}_{0:T}[k]$ \tcp*{Sec.\,\ref{sec:methods_hyperparameters}}
				
				Draw $\bi{\theta}[k] \: | \: \bi{z}_{0:T}[k], \bi{\vartheta}_{\mathrm{K}}[k]$ \tcp*{Sec.\,\ref{sec:methods_parameters}}
			}
		\end{algorithm}
	\end{minipage}

	\subsection{Closed-form Parameter Learning}\label{sec:methods_parameters}
	To construct the posterior of the paramters $\bi{\theta}$ from the estimated trajectories $\bi{z}_{0:T}$, we decompose
		$p(\bi{\theta}|\bi{z}_{0:T}, \bi{\vartheta}) \propto p(\bi{z}_{0:T} | \bi{\theta}, \bi{\vartheta}_{\mathrm{S}}) p(\bi{\theta} | \bi{\vartheta}_{\mathrm{K}})$
	using Bayes' rule. 
	Considering individual time steps, we can write the likelihood as
	\begin{equation}\label{eq:offline_parameter_decomposition}
		\begin{aligned}
			p(\bi{z}_{0:T} | \bi{\theta}, \bi{\vartheta}_{\mathrm{S}})  &= p(\bi{z}_0) \cdot \prod_{t=1}^{T} p(  \bi{z}_{t}  \mid  \bi{z}_{t-1}, \bi{\theta}, \bi{\vartheta}_{\mathrm{S}}) \propto \prod_{t=0}^{T}  p( \bi{h}_t  \mid \bi{x}_{t}, \bi{\theta}   ) \, , 
		\end{aligned}
	\end{equation}
	where the proportionality is regarding the parameters $\bi{\theta}$. 
	The factors $p( \bi{h}_t  \mid \bi{x}_{t}, \bi{\theta}   )$ are normal distributions, defined by \eqref{eq:probab_model2}, and the overall likelihood is, thus, a normal distribution
	\begin{equation}\label{eq:likelihood}
		\begin{aligned}
			p(\bi{z}_{0:T} | \bi{\theta}, \bi{\vartheta}_{\mathrm{S}})  & = \prod_{t=0}^{T} {b}_{t} \exp \left(  \sum_{i=1}^2 \bi{\alpha}_i^\top  \bi{s}_i(\bi{z}_{t}) -   \Tr \left( \bi{P}_i^\top(\bi{\alpha})  \bi{r}_i(\bi{x}_{t})\right) \right),
		\end{aligned}
	\end{equation}
	expressed in the canonical form of the restricted exponential family with the natural parameters {$\bi{\alpha}_1 =  \bi{a} / {\sigma}^{2}$} and $ \alpha_2 = - 1/(2{\sigma}^{2})$, respectively, log-partition functions $\bi{P}_i(\bi{\alpha})$, as well as (sufficient) statistics $\bi{s}_i (\bi{z}_{t})$ and $\bi{r}_i (\bi{x}_{t})$. 
	Please note that $\Tr(\cdot)$ is the trace operator, and that the base measures $b_t$ contain the $\bi{\theta}$-independent state dynamics $p(\bi{x}_{t} \mid  \bi{x}_{t-1}, \bi{h}_{t-1} , \bi{\vartheta}_{\mathrm{S}})$, defined by \eqref{eq:contin_model}. 
	Conveniently, the chosen $\mathcal{NIG}$ prior with distribution parameters $\bi{\eta} = \{\bi{0}, \bi{V}, {\psi}, \nu\}$, \ie
	\begin{equation}
		p(\bi{\theta} | \bi{\vartheta}_{\mathrm{K}}) = \mathcal{NIG}(\bi{a}, {\sigma}^2 \mid \bi{0}, \bi{V}, {\psi}, \nu)\\
		={n} (\bi{\eta}) \exp \left( \sum_{i=1}^2 \bi{\alpha}_i^\top \tilde{\bi{s}}_i(\bi{\eta})- \Tr \left( \bi{P}_i^\top(\bi{\alpha}) \tilde{\bi{r}}_i(\bi{\eta})\right)\right)\, ,
	\end{equation}
	is a conjugate prior to the likelihood \eqref{eq:likelihood}, with normalizing factor ${n} (\bi{\eta})$, as well as prior statistics $\tilde{\bi{s}}_i(\bi{\eta})$, and $\tilde{\bi{r}}_i(\bi{\eta})$, respectively.
	The parameter posterior $p(\bi{\theta}|\bi{z}_{0:T}, \bi{\vartheta})$ is available in closed form by summation of the prior statistics and the new statistics obtained from the estimates $\bi{z}_{0:T}$, that is $\bi{s}_i^+ = \tilde{\bi{s}}_i(\bi{\eta}) + \sum_{t=0}^T \bi{s}_i (\bi{z}_{t})$ and $\bi{r}_i^+ = \tilde{\bi{r}}_i(\bi{\eta}) + \sum_{t=0}^T \bi{r}_i (\bi{x}_{t})$. 
	The resulting posterior density is
	\begin{equation}\label{eq:offline_parameter_posterior}
		p(\bi{\theta}|\bi{z}_{0:T}, \bi{\vartheta}_{\mathrm{K}}) =\mathcal{NIG}(\bi{a}, {\sigma}^2 \mid \bi{m}^{+}, \bi{V}^{+}, \psi^{+}, \nu^{+}) \, ,
	\end{equation}
	with the posterior distribution parameters $\bi{\eta}^+ = \{\bi{m}^{+}, \bi{V}^{+}, \psi^{+}, \nu^{+}\}$ being
	\begin{equation} \label{eq:MdlStr_mniw_suffstat2para}
		\begin{aligned}
			\bi{m}^{+} &= \left(\bi{r}_1^+\right)^{-1} \bi{s}_{1}^+ \, , \qquad \bi{V}^{+} = \left(\bi{r}_1^+\right)^{-1} \, , \quad \; {\psi}^{+} = {s}_{2}^+ - \bi{s}_{1}^{+\top} \left(\bi{r}_1^+\right)^{-1} \bi{s}_{1}^+ \, ,  \quad \; \nu^{+} = r_2^+ \, .
		\end{aligned}
	\end{equation}
	The full derivation can be found in the Supplementary Material\,\ref{sup:restricted_exp_family}.
	
	\subsection{Hyperparameter Learning: Metropolis-within-Gibbs}\label{sec:methods_hyperparameters}
	For the hyperparameters $\bi{\vartheta} = \{ \bi{\vartheta}_{\mathrm{K}}, \bi{\vartheta}_{\mathrm{S}} \}$, similar closed-form results are not available. 
	Therefore, we employ Metropolis-within-Gibbs steps for learning, and, in each iteration $k$, generate a hyperparameter sample $\bi{\vartheta}^*$ from a proposal distribution $\zeta(\bi{\vartheta}^*|\bi{\vartheta}[k])$, \eg a random walk. 
	The proposals are accepted with probability $\min \left(1, \frac{p(\bi{\vartheta}^* | \bi{z}_{0:T})}{p(\bi{\vartheta}[k] | \bi{z}_{0:T})} \frac{\zeta(\bi{\vartheta}[k]|\bi{\vartheta}^*)}{\zeta(\bi{\vartheta}^*|\bi{\vartheta}[k])}\right)$, and rejected otherwise, \ie $\bi{\vartheta}[k+1] = \bi{\vartheta}[k]$. To evaluate $p(\bi{\vartheta}^* | \bi{z}_{0:T}) \propto p(\bi{z}_{0:T} | \bi{\vartheta}^*) p(\bi{\vartheta}^*)$, we use the hyperparameter prior $p(\bi{\vartheta}^*)$ and compute the likelihood $p(\bi{z}_{0:T} | \bi{\vartheta}^*)$ for the kernel and structural hyperparameters individually.
	
	\paragraph{Kernel Hyperparameters}\label{sec:kernel_hyperparam}
	For the likelihood of the \ac{gp} kernel hyperparameters $\bi{\vartheta}_{\mathrm{K}}$, we have
	\begin{equation}\label{eq:data_likelihood_hyperparam}
		p(\bi{z}_{0:T} | \bi{\vartheta}_{\mathrm{K}}^*) = \frac{p(\bi{\theta} | \bi{\vartheta}_{\mathrm{K}}^*) p(\bi{z}_{0:T} | \bi{\theta}, \bi{\vartheta}_{\mathrm{K}}^*) }{p( \bi{\theta} | \bi{z}_{0:T}, \bi{\vartheta}_{\mathrm{K}}^*)} = \frac{n(\bi{\eta}^{*}) \prod_{t=0}^{T} b_t  }{n(\bi{\eta}^{+*})} \, ,
	\end{equation}
	using Bayes' rule, and we note that all terms are known in closed form. 
	The numerator and the denominator in \eqref{eq:data_likelihood_hyperparam} are proportional to each other. 
	In fact, both follow a $\mathcal{NIG}$ distribution, and the $\bi{\theta}$-related components cancel. 
	Therefore, the likelihood for the kernel hyperparameters is a quotient of normalizing constants, 
	and the first term of the acceptance probability can be computed as 
	\begin{equation}
		\begin{aligned}
			\frac{p(\bi{\vartheta}_{\mathrm{K}}^* | \bi{z}_{0:T})}{p(\bi{\vartheta}_{\mathrm{K}}[k] | \bi{z}_{0:T})}  &= \frac{p(\bi{z}_{0:T} | \bi{\vartheta}_{\mathrm{K}}^*) p(\bi{\vartheta}_{\mathrm{K}}^*)}{p(\bi{z}_{0:T} | \bi{\vartheta}_{\mathrm{K}}[k]) p(\bi{\vartheta}_{\mathrm{K}}[k])} = \frac{ n(\bi{\eta}^{*}) p(\bi{\vartheta}_{\mathrm{K}}^*)}{n(\bi{\eta}^{+*})} \frac{n(\bi{\eta}^{+}[k])}{ n(\bi{\eta}[k]) p(\bi{\vartheta}_{\mathrm{K}}[k])} \, ,
		\end{aligned}
	\end{equation}
	where the base measure products $\prod_{t=0}^{T} b_t$ in the numerator and the denominator cancel.
	
	\paragraph{Structural Hyperparameters}\label{sec:structural_hyperparam}
	For the structural hyperparameters $\bi{\vartheta}_{\mathrm{S}}$, the likelihood can be constructed by integrating out the parameters $\bi{\theta}$ from $p(\bi{z}_{0:T}, \bi{\theta}|  \bi{\vartheta}_{\mathrm{S}})$. 
    This is done by noting that the result is proportional regarding $\bi{\vartheta}_{\mathrm{S}}$ to the Gaussian density $p(\bi{x}_{0:T} \mid  \bi{h}_{0:T-1}, \bi{\vartheta}_{\mathrm{S}})$, that is
	\begin{equation}
		\begin{aligned}
			p(\bi{z}_{0:T} | \bi{\vartheta}_{\mathrm{S}} ) &=  \int \left[ p(\bi{x}_0) \cdot \prod_{t=0}^{T} p( \bi{h}_t  \mid \bi{x}_{t}, \bi{\theta}   ) \prod_{t=1}^{T} p(\bi{x}_{t} \mid  \bi{x}_{t-1}, \bi{h}_{t-1} , \bi{\vartheta}_{\mathrm{S}}) p(\bi{\theta}) \right] \, \mathrm{d}\bi{\theta} \\
			& \propto  p(\bi{x}_0) \cdot \prod_{t=1}^{T} p(\bi{x}_{t} \mid \bi{x}_{t-1}, \bi{h}_{t-1} , \bi{\vartheta}_{\mathrm{S}}) = p(\bi{x}_{0:T} \mid    \bi{h}_{0:T-1}, \bi{\vartheta}_{\mathrm{S}}) \, .
		\end{aligned}
	\end{equation}

    Thus, we compute the first quotient of the acceptance probability as
	\begin{equation}
		\begin{aligned}
			\frac{p(\bi{\vartheta}_{\mathrm{S}}^* | \bi{z}_{0:T})}{p(\bi{\vartheta}_{\mathrm{S}}[k] | \bi{z}_{0:T})}  &=  \frac{p(\bi{x}_{0:T} \mid   \bi{h}_{0:T-1}, \bi{\vartheta}_{\mathrm{S}}^*) p(\bi{\vartheta}_{\mathrm{S}}^*)}{p(\bi{x}_{0:T} \mid    \bi{h}_{0:T-1}, \bi{\vartheta}_{\mathrm{S}}[k]) p(\bi{\vartheta}_{\mathrm{S}}[k])}\, ,
		\end{aligned}
	\end{equation}
	where the normalizing constants cancel.

	\begin{wrapfigure}{r}{0.4\textwidth} 
		\vspace{-20mm}
		\centering
		{\fontsize{7pt}{7pt}\selectfont
			\resizebox{0.4\textwidth}{!}{
				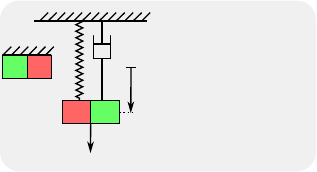
			}
		}
		\vspace{-6mm}
		\caption{Test system \& Hamiltonian.}
		\label{fig:problem_setting}
        \vspace{-5mm}
	\end{wrapfigure}
	
	\section{Simulation and Results}\label{sec:results}
	To evaluate the proposed method, we conduct a simulation case study with a non-harmonic oscillator, governed by the Hamiltonian $H(q,p) =  \frac{q^2}{2} +   \frac{p^2}{2}  +   2  \cos q  $, with the position $q$ and the momentum $p$ (see Figure\,\ref{fig:problem_setting}).  
	The system is driven by a known input force $u$ and dissipates energy through a damping coefficient $d =: \vartheta_{\mathrm{S}}$, to be estimated. 
	To generate training data, we simulate a single trajectory by applying an input signal $u_t$ for time steps $t=0,\hdots,T$, and storing only noisy position measurements $y^{(\mathrm{i/o})}_t = q_t + e_t$ as outputs.\footnote{Further details on the simulation (including the matrices $\bi{J}, \bi{R}$, $\bi{G}$), the algorithmic setup, as well as the training/ testing scenarios can be found in the Suppl. Material\,\ref{sup:simulation_setup}. Convergence metrics are given in Suppl. Material \ref{sup:comp_metrics}.}
	For comparison with existing work that relies on full-state measurements, we also consider an input-state setting, \ie $\bi{y}^{(\mathrm{i/s})}_t = [q_t, p_t]^\top + \bi{e}_t$. 
	In particular, we compare the proposed Algorithm\,\ref{alg:Bayesian_inference_learning} for physically consistent learning from input-output data to \textit{(i)} itself in an input-state setting, and to \textit{(ii)} an exact Hamiltonian \ac{gp} model based on \cite{Beckers.2022}. 
	Please note that we set non-informative priors on all parameters, for instance, broad Gaussian distributions for the hyper-prior $p(\bi{\vartheta})$.

	In Figure\,\ref{fig:Hamiltonians}, the flow map resulting from the true Hamiltonian (left) 
	is compared to the learned Hamiltonian models in terms of their \ac{rmse} and computational complexity. 
	While the algorithms in the middle columns have access to state measurements, the proposed approach (right) uses only noisy position measurements, \ie input-output data. 
	Despite only having access to input-output data, the proposed method learns the flow map with accuracy comparable to approaches that have access to full-state measurements. 
    Please note that learning the considered non-harmonic oscillator system from input-output data required incorporating a symmetry constraint.\footnote{\label{footn} Due to identifiability issues in the input-output setting, we impose a symmetry constraint on the Hamiltonian \ac{gp}, \ie $\widehat{H}(\bi{x}) = \widehat{H}(-\bi{x})$, which induces an anti-symmetric gradient $\nablabx \widehat{H}(\bi{x}) = - \nablabx \widehat{H}(-\bi{x})$. Results without symmetry constraint are presented in the Suppl. Material\,\ref{sup:wo_symmetry_sontraint}. No symmetry constraint is used in the input-state setting.}
    In this regard, future work may examine conditions for the Hamiltonian's identifiability.

	\newcommand{\figwidth}{0.23\textwidth}
	\newcommand{\figwidthnew}{0.243\textwidth}
	\newcommand{\myfont}[1]{\textbf{\fontsize{8pt}{8pt}\selectfont #1}}

	\begin{figure}[tb]
		\centering
		\newcolumntype{C}[1]{>{\centering\arraybackslash}p{#1}}
		
		{\fontsize{7pt}{7pt}\selectfont
			\begin{tabular}{p{0.225\textwidth} | C{0.225\textwidth}  C{0.225\textwidth} C{0.225\textwidth}}
				\toprule[1pt]
				\myfont{Hamiltonian model} & \myfont{Exact \acs{gp}} (\smallsquare{0.7843}{0.8275}{ 0.0902}) & \myfont{Reduced-rank \acs{gp}} (\smallsquareline{0.9059}{0.4824}{0.1608})   & \myfont{Reduced-rank \acs{gp}}\hyperref[footn]{$^4$} (\smallsquareline{0}{0.3137}{0.6078}) \\
				\midrule
				Complexity (mean prediction) & $\mathcal{O}(T)$ & $\mathcal{O}(M)$ & $\mathcal{O}(M)$ \\
				Training data & \textcolor[RGB]{120,0,0}{input-state data} & \textcolor[RGB]{120,0,0}{input-state data} & \textcolor[RGB]{0,120,0}{input-output data} \\
				Flow magnitude \ac{rmse} & $3.84\,\mathrm{J}$ & $3.09\,\mathrm{J}$ & $3.27\,\mathrm{J}$ \\
				Flow angle \ac{rmse}  & $0.04\,\mathrm{rad}$ & $0.08\,\mathrm{rad}$ & $0.12\,\mathrm{rad}$ \\ 
				\midrule
				\resizebox{\figwidthnew}{!}{
\begingroup%
  \makeatletter%
  \providecommand\color[2][]{%
    \errmessage{(Inkscape) Color is used for the text in Inkscape, but the package 'color.sty' is not loaded}%
    \renewcommand\color[2][]{}%
  }%
  \providecommand\transparent[1]{%
    \errmessage{(Inkscape) Transparency is used (non-zero) for the text in Inkscape, but the package 'transparent.sty' is not loaded}%
    \renewcommand\transparent[1]{}%
  }%
  \providecommand\rotatebox[2]{#2}%
  \newcommand*\fsize{\dimexpr\f@size pt\relax}%
  \newcommand*\lineheight[1]{\fontsize{\fsize}{#1\fsize}\selectfont}%
  \ifx\svgwidth\undefined%
    \setlength{\unitlength}{106bp}%
    \ifx\svgscale\undefined%
      \relax%
    \else%
      \setlength{\unitlength}{\unitlength * \real{\svgscale}}%
    \fi%
  \else%
    \setlength{\unitlength}{\svgwidth}%
  \fi%
  \global\let\svgwidth\undefined%
  \global\let\svgscale\undefined%
  \makeatother%
  \begin{picture}(1,0.94339623)%
    \lineheight{1}%
    \setlength\tabcolsep{0pt}%
    \put(0,0){\includegraphics[width=\unitlength,page=1]{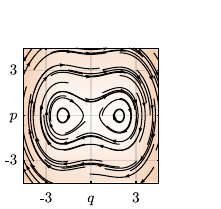}}%
    \put(0.80980367,0.10805872){\color[rgb]{0,0,0}\rotatebox{90}{\makebox(0,0)[lt]{\lineheight{1.10000002}\smash{\begin{tabular}[t]{l}Flow magnitude in $\mathrm{J}$\end{tabular}}}}}%
    \put(0.00161812,0.88645711){\color[rgb]{0,0,0}\makebox(0,0)[lt]{\lineheight{1.10000002}\smash{\begin{tabular}[t]{l}Hamiltonian gradient flow\\True vs. learned (right)\end{tabular}}}}%
    \put(0,0){\includegraphics[width=\unitlength,page=2]{Fig1_vp_true_annotated.pdf}}%
    \put(0.88527269,0.70422714){\color[rgb]{0,0,0}\makebox(0,0)[lt]{\lineheight{1.14999998}\smash{\begin{tabular}[t]{l}20\\\\15\\\\10\\\\5\\\\0\end{tabular}}}}%
    \put(0,0){\includegraphics[width=\unitlength,page=3]{Fig1_vp_true_annotated.pdf}}%
  \end{picture}%
\endgroup%
} &
				\includegraphics[width=\figwidthnew]{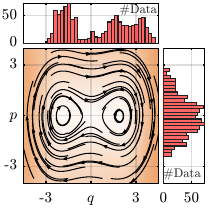} &
				\includegraphics[width=\figwidthnew]{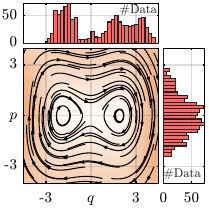} &
				\includegraphics[width=\figwidthnew]{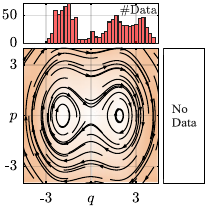} \\
				\bottomrule[1pt]
		\end{tabular}}

		\caption{Flow maps following from the true and learned Hamiltonians. The reduced-rank Hamiltonian \ac{gp} yields similar approximation accuracy while allowing for computationally efficient prediction. Despite only having access to input-output data, the proposed method (right) enables learning a Hamiltonian \ac{gp} with accuracy comparable to methods with access to full-state measurements.}
		\label{fig:Hamiltonians}
	\end{figure}

	\begin{figure}[tb]
		\centering
        \begin{minipage}[t]{0.645\textwidth}
            \includegraphics[width=1\textwidth]{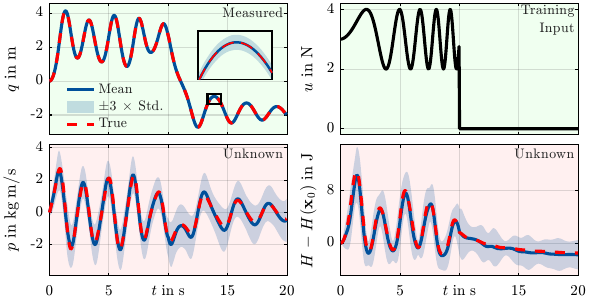}
        \end{minipage}
		\begin{minipage}[t]{0.345\textwidth}
            \includegraphics[width=1\textwidth]{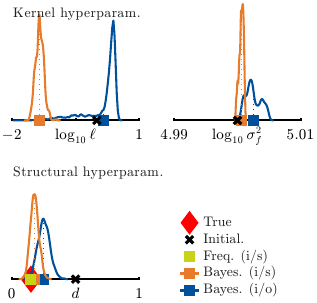}
        \end{minipage}

		\caption{True and estimated system behavior in the training data set (left), as well as density estimates for the \ac{gp} kernel and structural hyperparameters $\bi{\vartheta} = \{ \bi{\vartheta}_{\mathrm{K}}, {\vartheta}_{\mathrm{S}} \}$ (right). From input-output data, the proposed method infers densities of unknown hidden states, the current system energy, and hyperparameters in a fully Bayesian fashion.}
		\label{fig:Trajectories_train}
	\end{figure}

	Figure\,\ref{fig:Trajectories_train} depicts the Bayesian inference and learning results on the training data set, showcasing that the method provides density estimates for the model (hyper-)parameters, and accurate smoothing estimates for all state variables (including unknown hidden states). 
	If we run Algorithm\,\ref{alg:Bayesian_inference_learning} with measurements $\bi{y}^{(\mathrm{i/s})}_{0:T}$, the hyperparameter density estimates are pronounced, and the density $p(d|\bi{y}^{(\mathrm{i/s})}_{0:T})$ accurately reflects the true damping coefficient. 
	In comparison, the density $p(d|{y}^{(\mathrm{i/o})}_{0:T})$ is slightly biased, which we attribute to identifiability issues in the chosen simulation example.

	To evaluate the physical consistency of the learned Hamiltonian \ac{gp}, we draw models from the sample-based posterior distribution, provided by Algorithm\,\ref{alg:Bayesian_inference_learning}, and compare their forward simulations from an initial value with a true system trajectory (see Figure\,\ref{fig:Trajectories_pred} and details in the Supplementary Material\,\ref{sup:simulation_setup}). 
	Despite learning from input-output data, Algorithm\,\ref{alg:Bayesian_inference_learning} yields a physically consistent, probabilistic model whose samples closely resemble the actual system behavior. 
	In fact, the method can represent multimodal densities in phase space, as can be seen in a single sample converging to the system's upper equilibrium (compare Figure\,\ref{fig:problem_setting} and Figure\,\ref{fig:Trajectories_pred}, top left). 
	Notably, the method consistently accounts for dissipation and energy intake through exogenous inputs, as visible in the monotonically decreasing Hamiltonian $H$ after the input signal reaches zero.

		\section{Conclusion}\label{sec:conclusion}
		In this article, we present a method for learning a physically consistent dynamics model using non-conservative Hamiltonian \acp{gp}.
		In contrast to existing work, which relies on measurements of momenta or velocities, we consider learning solely based on input-output data. 
		The proposed reduced-rank Hamiltonian \ac{gp} is linear in its parameters and comes with closed-form gradient expressions. 
		While we exploit these properties for fully Bayesian learning of \mbox{(hyper-)}parameters, we emphasize that the model structure can also speed up training and prediction \textit{significantly} in frequentist settings. 
		However, we acknowledge that the computational complexity of the particle-based scheme and the reduced-rank \ac{gp} increases rapidly with the problem dimension \citep{RiutortMayol.2023}, which future work should consider. 
        Moreover, the presented empirical case study necessitated prior knowledge in the form of symmetry constraints to learn a complex Hamiltonian function, highlighting the relevance of a future identifiability analysis.

		Taking a step back, we present an approach to bridge physics-informed dynamics learning \citep{Beckers.2022} and physically consistent Bayesian system identification. 
		Using the proposed method, an uncertainty-quantifying, physically consistent nonlinear system model can be learned solely from input-output data, potentially enabling model-based control and insights in complex application systems without extensive prior knowledge.

		\newpage
		\acks{This research was supported by the \textit{Kjell och Märta Beijer Foundation} and the \textit{Swedish Research Council (VR)} under the contract numbers 2021-04321 and 2025-04318. Jan-Hendrik Ewering was supported by the \textit{German Academic Scholarship Foundation (Studienstiftung des Deutschen Volkes)}.}
		
		\bibliography{l4dc_2026}
		
		\clearpage

		\renewcommand{\thesection}{\Alph{section}} 
		\setcounter{section}{0} 

		\section*{Supplementary Material}
		
		\section{Proposed Model in the Canonical Form of the Restricted Exponential Family}\label{sup:restricted_exp_family}
		
		To derive the distributions employed in Section\,\ref{sec:inference_learning}, we express all densities in the canonical form of the restricted exponential family. 
		This is convenient for Bayesian inference in state-space models, as probabilistic dependence on the previous state can be incorporated \citep{Wigren.2019}. 
		
		\paragraph{Formulation}
		The proposed model \eqref{eq:contin_model} can be written in the canonical form of the restricted exponential family as
		\begin{equation}\label{eq:canonical_form}
			p(\bi{\alpha}) =  Z \exp \left( \sum_{i=1}^2 \bi{\alpha}_i^\top {\bi{s}}_i- \Tr \left( \bi{P}_i^\top(\bi{\alpha}) {\bi{r}}_i\right)\right)\, ,
		\end{equation}
		where the density is expressed in terms of the \textit{natural parameters} $\bi{\alpha}$ that are obtained by transforming the original model parameters $\bi{\theta} = \{\bi{a}, \sigma^2\}$. 
		We have $\bi{\alpha} = [\bi{\alpha}_1^\top, \alpha_2]^\top$ with
		\begin{equation}
			\bi{\alpha}_1 =  \frac{\bi{a}}{{\sigma}^{2}} \, , \qquad \alpha_2 = - \frac{1}{2{\sigma}^{2}}  \, ,
		\end{equation}
		and the parameter-dependent parts of the log-partition function being
		\begin{align}
			\bi{P}_1 (\bi{\alpha}) &= -\frac{1}{4} \bi{\alpha}_1 {\alpha}_2^{-1}\bi{\alpha}_1^\top = \frac{1}{2{\sigma}^{2}} \bi{a} \bi{a}^\top \, , \qquad {P}_2 (\bi{\alpha}) =  -\frac{1}{2} \log |-2{\alpha}_2 |=
			\frac{1}{2} \log {\sigma}^2   \, ,
		\end{align}
		where we denote $|\cdot|$ as the determinant. 
		The remaining variables, $Z$, $\bi{s}_i$, and $\bi{r}_i$---depending on the employed prior and the data---are a normalization constant and statistics, respectively.
		
		\paragraph{Likelihood}
		To express the likelihood
		\begin{equation}\label{eq:offline_parameter_decomposition_sup}
			\begin{aligned}
				p(\bi{z}_{0:T} | \bi{\theta}, \bi{\vartheta}_{\mathrm{S}})        &= p(\bi{z}_0) \cdot \prod_{t=1}^{T} p(  \bi{z}_{t}  \mid  \bi{z}_{t-1}, \bi{\theta}, \bi{\vartheta}_{\mathrm{S}}) \\
				&= p(\bi{x}_0) p( \bi{h}_0  \mid \bi{x}_{0}, \bi{\theta}   ) \cdot \prod_{t=1}^{T} p(  \bi{z}_{t}  \mid  \bi{z}_{t-1}, \bi{\theta}, \bi{\vartheta}_{\mathrm{S}}) \,,
			\end{aligned}
		\end{equation}
		in the form \eqref{eq:canonical_form}, we assume, for simplicity, that $p(\bi{x}_0)$ is known and consider each time step separately. 
		For individual transitions \mbox{$p(\bi{z}_{t}\mid\bi{z}_{t-1},\bi{\theta},\bi{\vartheta}_{\mathrm{S}})$} we have
		\begin{equation}
			\begin{aligned}
				p(  \bi{z}_{t}  \mid  \bi{z}_{t-1}, \bi{\theta}, \bi{\vartheta}_{\mathrm{S}}) &=  p( \bi{h}_t  \mid \bi{x}_{t}, \bi{\theta}   ) \, p(\bi{x}_{t} \mid  \bi{x}_{t-1}, \bi{h}_{t-1} , \bi{\vartheta}_{\mathrm{S}}) \\
				&= \mathcal{N} (\bi{h}_t  \mid \bi{D}_{\bi{\phi}}(\bi{x}_t) \bi{a}, {\sigma^2} \mathbf{I}) \, \mathcal{N} ({\bi{x}}_t \mid \bi{f}({\bi{x}}_{t-1}, \bi{h}_{t-1} , \bi{u}_{t-1}, \bi{\vartheta}_{\mathrm{S}}) , \bi{\Sigma}_{w}) \, ,
			\end{aligned}
		\end{equation}
		were the function $\bi{f}$ and the noise covariance $\bi{\Sigma}_w$ describe the discrete-time state dynamics, and can be obtained from integrating \eqref{eq:contin_model}. 
		Noting that only the first density is $\bi{\theta}$-dependent, the likelihood for one time step can be expressed as
		\begin{equation}
			\begin{aligned}
				& p(  \bi{z}_{t}  \mid  \bi{z}_{t-1}, \bi{\theta}, \bi{\vartheta}_{\mathrm{S}}) = \tilde{b}_t \cdot \mathcal{N} (\bi{h}_t  \mid \bi{D}_{\bi{\phi}}(\bi{x}_t) \bi{a}, {\sigma^2} \mathbf{I})  \\
				&\qquad \qquad = \frac{\tilde{b}_t}{(2 \pi)^{n_x / 2}({\sigma^2})^{n_x / 2}} \exp \left(-\frac{1}{2{\sigma}^{2}} (\bi{h}_t  -\bi{D}_{\bi{\phi}}(\bi{x}_t) \bi{a})^\top (\bi{h}_t -\bi{D}_{\bi{\phi}}(\bi{x}_t) \bi{a})\right) \\
				& \qquad \qquad = {b}_t \exp \left( \sum_{i=1}^{2}  \bi{\alpha}_i^\top  \bi{s}_i ( \bi{z}_t) -  \Tr \left(\bi{P}_i^\top (\bi{\alpha})  \bi{r}_i (\bi{x}_t) \right)\right) \, ,
			\end{aligned}
		\end{equation}
		with the statistics
		\begin{equation}\label{eq:statistics_sup}
			\begin{aligned}
				\bi{s}_1 (\bi{z}_{t}) & =		    \bi{D}_{\bi{\phi}}\left(\bi{x}_t\right)^{\top} \bi{h}_t  \, , \qquad & {s}_2 (\bi{z}_{t}) & =             \bi{h}_t^{\top} \bi{h}_t \, , \\
				\bi{r}_1 (\bi{x}_{t}) & = \bi{D}_{\bi{\phi}}\left(\bi{x}_t\right)^{\top} \bi{D}_{\bi{\phi}}\left(\bi{x}_t\right)\, , \qquad & r_2 (\bi{x}_{t}) & = n_x \, ,
			\end{aligned}
		\end{equation} 
		and the base measures
		\begin{equation}
			{b}_t =  \begin{cases}
				{(2 \pi)^{- n_x / 2}} \cdot p(\bi{x}_0) \, , & \text{if  } t=0 \, ,\\
				{(2 \pi)^{- n_x / 2}} \cdot \mathcal{N} ({\bi{x}}_t \mid \bi{f}({\bi{x}}_{t-1}, \bi{h}_{t-1} , \bi{u}_{t-1}, \bi{\vartheta}_{\mathrm{S}}) , \bi{\Sigma}_{w}) \,, & \text{otherwise}\, . 
			\end{cases} 
		\end{equation}
		The overall likelihood for data $\bi{z}_{0:T}$ is constructed by multiplying the one-step densities, which amounts to summing the corresponding trajectory statistics \eqref{eq:statistics_sup}.

		\paragraph{Prior and Posterior} 
		To express the multivariate normal inverse Gamma ($\mathcal{NIG}$) \citep{Murphy.2007} prior and posterior densities in the form \eqref{eq:canonical_form}, we first define the 
		normal distribution 
		\begin{equation}
			\begin{aligned}
				p(\bi{a} \mid \bi{m}, \bi{V}, {\sigma^2})&= \mathcal{N} (\bi{a} \mid \bi{m}, {\sigma^2} \bi{V}) \\
				&= \frac{1}{(2 \pi)^{M / 2}({\sigma^2})^{M / 2}|\bi{V}|^{1 / 2}} \exp \left(-\frac{1}{2{\sigma}^{2}} (\bi{a}-\bi{m})^\top \bi{V}^{-1}(\bi{a}-\bi{m})\right) \, ,
			\end{aligned}
		\end{equation}
		and inverse Gamma ($\mathcal{IG}$) distribution
		\begin{align}
			p({\sigma^2} \mid {\psi}, \nu) = \mathcal{IG}({\sigma^2} \mid {\psi}, \nu) = \frac{({\psi} /2 )^{\nu / 2}}{ \Gamma\left(\frac{\nu}{2}\right)}({\sigma^2})^{-\nu/ 2 - 1 } \exp \left( -\frac{1}{2 {\sigma}^{2}} {\psi}  \right) \, ,
		\end{align}
		where $\Gamma$ is the scalar Gamma function. 
		Given this, the prior and posterior densities can be expressed in the canonical form of the restricted exponential family as
		\begin{equation}\label{eq:prior_canonical}
			\begin{aligned}
				&p(\bi{a}, \sigma^2 \mid \bi{m}, \bi{V}, {\psi}, \nu)= \mathcal{NIG} (\bi{a} \mid \bi{m},\bi{V}, {\psi}, \nu) \\
				& \qquad \qquad = \frac{({\psi} /2 )^{\nu / 2} \left(\frac{1}{\sigma^2}\right)^{\nu/ 2 + 1 + M/2}}{(2 \pi)^{M / 2}|\bi{V}|^{1 / 2} \Gamma\left(\frac{\nu}{2}\right)}  \exp \left(-\frac{1}{2{\sigma}^{2}} \left[ \psi + (\bi{a}-\bi{m})^\top \bi{V}^{-1}(\bi{a}-\bi{m}) \right] \right) \\
				& \qquad \qquad ={n} (\bi{\eta}) \exp \left( \sum_{i=1}^{2} \bi{\alpha}_i^\top  \tilde{\bi{s}}_i(\bi{\eta}) -  \Tr \left( \bi{P}_i^\top(\bi{\alpha}  )  \tilde{\bi{r}}_i(\bi{\eta}) \right)\right)\, ,
			\end{aligned}
		\end{equation}
		where the statistics, dependent on the distribution parameters $\bi{\eta} = \{\bi{m}, \bi{V}, {\psi}, \nu \}$, are
		\begin{equation}
			\begin{aligned}
				\tilde{\bi{s}}_1 (\bi{\eta}) & = 
				\bi{V}^{-1}\bi{m}\,, \qquad & \tilde{{s}}_2 (\bi{\eta}) & =            \psi + \bi{m}^\top \bi{V}^{-1}\bi{m} \, , \\
				\tilde{\bi{r}}_1 (\bi{\eta}) & =\bi{V}^{-1}\, , \qquad & \tilde{r}_2 (\bi{\eta}) & = \nu + 2 + M \, ,
			\end{aligned}
		\end{equation} 
		and the normalizing factor 
		\begin{equation}
			{n}(\bi{\eta}) = \frac{({\psi} /2 )^{\nu / 2}}{(2 \pi)^{M / 2}|\bi{V}|^{1 / 2} \Gamma\left(\frac{\nu}{2}\right)} \, .
		\end{equation}
		
		The parameter posterior $p(\bi{\theta}|\bi{z}_{0:T}, \bi{\vartheta})$ is available in closed form by summation of the prior statistics and new statistics, obtained from the data trajectories $\bi{z}_{0:T}$, that is $\bi{s}_i^+ = \tilde{\bi{s}}_i(\bi{\eta}) + \sum_{t=0}^T \bi{s}_i (\bi{z}_{t})$ and $\bi{r}_i^+ = \tilde{\bi{r}}_i(\bi{\eta}) + \sum_{t=0}^T \bi{r}_i (\bi{x}_{t})$. 
		The resulting parameter posterior density is
		\begin{equation}\label{eq:offline_parameter_posterior2}
			p(\bi{\theta}|\bi{z}_{0:T}, \bi{\vartheta}_{\mathrm{K}}) =\mathcal{NIG}(\bi{a}, {\sigma}^2| \bi{m}^{+}, \bi{V}^{+}, \psi^{+}, \nu^{+}) \, ,
		\end{equation}
		with the new distribution parameters $\bi{\eta}^+ = \{\bi{m}^{+}, \bi{V}^{+}, \psi^{+}, \nu^{+}\}$ being
		\begin{equation} \label{eq:MdlStr_mniw_suffstat2para2}
			\begin{aligned}
				\bi{m}^{+} &= \left(\bi{r}_1^+\right)^{-1} \bi{s}_{1}^+ \, , \qquad & \bi{V}^{+} &= \left(\bi{r}_1^+\right)^{-1} \, ,\\
				{\psi}^{+} &= {s}_{2}^+ - \bi{s}_{1}^{+\top} \left(\bi{r}_1^+\right)^{-1} \bi{s}_{1}^+ \, , \qquad & \nu^{+} &= r_2^+ \, .
			\end{aligned}
		\end{equation}

		\newpage
		\section{Simulation Case Study}\label{sup:simulation_setup}
		
		\paragraph{Simulation Setup} 
		We conduct a simulation case study with a non-harmonic oscillator, governed by the Hamiltonian function
		\begin{equation}
			H \left( \begin{bmatrix}
				q \\ p
			\end{bmatrix}\right) =  \frac{q^2}{2} +   \frac{p^2}{2}  +   2  \cos q \, ,
		\end{equation}
		with position $q$ and momentum $p$ (see Figure\,\ref{fig:problem_setting}).  
		The system is driven by a known input force $u$ and dissipates energy through a damping coefficient $d$, to be estimated. 
		We consider the system 
		\begin{equation}
			\begin{aligned}
				\frac{\mathrm{d}}{\mathrm{d}t} \bi{x} &= \frac{\mathrm{d}}{\mathrm{d}t} \begin{bmatrix}
					q \\ p
				\end{bmatrix} = \left( \underbrace{\begin{bmatrix} 0 & 1 \\ -1 & 0\end{bmatrix}}_{\bi{J}} - \underbrace{\begin{bmatrix}
						0 & 0 \\ 0 &  d
				\end{bmatrix}}_{\bi{R}}\right) \nablabx H(\bi{x}) + \underbrace{\begin{bmatrix}
						0 \\ 1
				\end{bmatrix}}_{\bi{G}} {u} + \bi{w} \, , \\
				\bi{y}^{(\mathrm{i/s})} &= \bi{x} + \bi{e}^{(\mathrm{i/s})} \, , \\
				{y}^{(\mathrm{i/o})} &= q + e^{(\mathrm{i/o})} \, ,
			\end{aligned}
		\end{equation}
		with damping coefficient $d=0.15$. The zero-mean discrete-time process and measurement noise terms are Gaussian with standard deviation $10^{-4}$ and $10^{-3}$, respectively. 
		To generate training data, we simulate the system from the initial value $\bi{x}_0 = [0,0]^\top$, using the input signal illustrated in Figure\,\ref{fig:Trajectories_train}, for $T=1,000$ time steps with a symplectic Euler integrator at step size $\delta = 0.02\,\mathrm{s}$.

		\paragraph{Inference and Learning}
		To perform Bayesian inference and learning, we run Algorithm\,\ref{alg:Bayesian_inference_learning} for \mbox{$K = 20,000$} iterations. 
		We discard the first $10,000$ samples to ignore the burn-in period of the Markov chain. 
		For discretizing the state dynamics \eqref{eq:contin_model} in Algorithm\,\ref{alg:Bayesian_inference_learning}, we use Euler integration. 
		To generate proposals in the Metropolis-within-Gibbs steps, a random walk is employed for the kernel hyperparameters $\bi{\vartheta}_{\mathrm{K}}$. 
		For the structural hyperparameter ${\vartheta}_{\mathrm{S}} := d$, we use refined proposals based on the gradient and Hessian of the likelihood \citep{Roberts.2002}. 
		The chosen parameters are summarized in Table\,\ref{tab:parameters}. 
		
		\begin{table}[h]
			\centering
			
			\caption{Parameters of Algorithm\,\ref{alg:Bayesian_inference_learning} in the Case Study}
			{\fontsize{9pt}{9pt}\selectfont
				\begin{tabular}{lll}
					\toprule[1pt]
					Name  & Symbol & Value \\
					\midrule
					Number of eigenfunctions & $M$   & 20 without symmetry constraint \\
					&    & 15 with symmetry constraint \\
					Domain bounds & $L_1, L_2$ & 8, 8 \\
					Hyper-prior  & $p(\bi{\vartheta})$ &near uniform (broad Gaussian) \\
					Scale & $\psi$ &  100 \\
					Degrees of freedom & $\nu$ & 400 \\
					\bottomrule[1pt]
				\end{tabular}%
			}
			\label{tab:parameters}%
		\end{table}%

\begin{figure}[H] 
    		\centering
    		\includegraphics[width=1\textwidth]{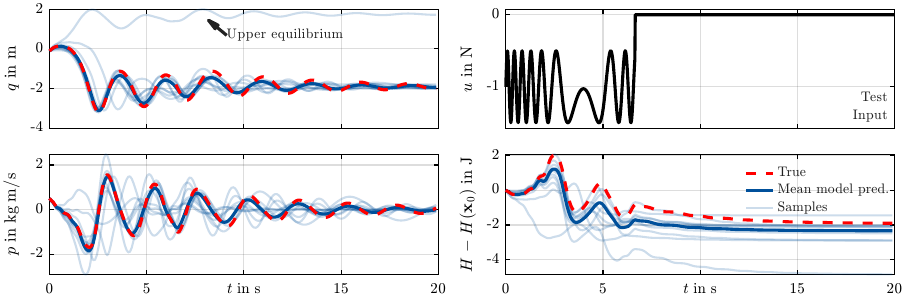}
    		\caption{True system behavior and forward predictions of the learned Hamiltonian \ac{gp} model. 
    			The proposed method provides a probabilistic Hamiltonian system, and---despite learning only from input-output data---\textit{each} sampled model yields a physically consistent prediction that resembles the actual system behavior.}
    		\label{fig:Trajectories_pred}
    	\end{figure}

		\paragraph{Testing}
		For testing in Figure\,\ref{fig:Trajectories_pred}, we use the samples $\{ \bi{\theta}[k], \vartheta_{\mathrm{S}}[k] \}$, $k = 10,000, \hdots, 20,000$, provided by Algorithm\,\ref{alg:Bayesian_inference_learning} to \textit{(i)} generate a mean model and \textit{(ii)} draw ten random instances, each representing a sample from the posterior model distribution. 
		Using these models, we perform forward-predictions with a symplectic Euler integration scheme at step size $\delta = 0.01\,\mathrm{s}$. We simulate from the initial value $\bi{x}_0 = [-0.1, 0.5]^\top$ and apply a pre-set test input signal (depicted in Figure\,\ref{fig:Trajectories_pred}), different from the training input signal. 

        \clearpage
		\section{Convergence Metrics}\label{sup:comp_metrics}
\begin{figure}[h] 
    		\centering
    		\includegraphics[width=1\textwidth]{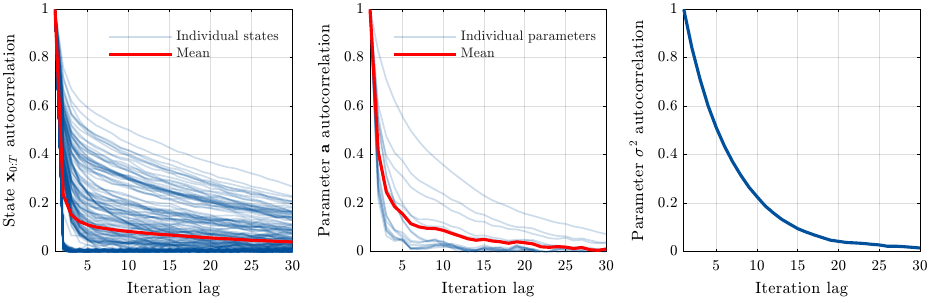}
    		\caption{Autocorrelation of the state and parameter samples. In the plots, each state time step and each state/parameter dimension are represented by individual autocorrelation graphs.}
    		\label{fig:metrics_1}
    	\end{figure}

        For the simulation case study in Section\,\ref{sec:results}, we evaluate the convergence of the \ac{pmcmc} scheme in Algorithm\,\ref{alg:Bayesian_inference_learning}. 
        Specifically, we present the autocorrelation plots of individual states and parameters in Figure\,\ref{fig:metrics_1}, which indicate the degree of correlation between successive draws from the sampler. 
        A quickly decreasing autocorrelation indicates good ``mixing'' and, thus, an efficient exploration of the domain \citep{Wigren.2022}. 
        
\begin{figure}[h] 
    		\centering
    		\includegraphics[width=1\textwidth]{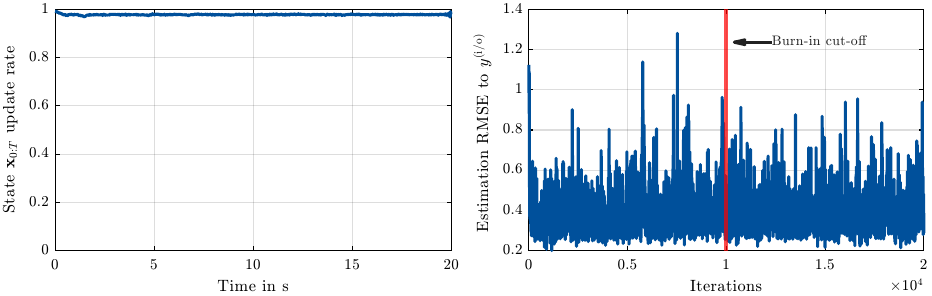}
    		\caption{Update rate of the state samples (left) and \ac{rmse} between the estimated positions and the true position measurements $y^{\mathrm{(i/o)}}$ (right).}
    		\label{fig:metrics_2}
    	\end{figure}

        Similarly, the update rate along the time axis provides a measure of how regularly state time step samples $\bi{x}_t [k]$ are updated throughout the sampling procedure. 
        Ideally, the update rate should be close to $1$ for each time step, which is the case in the present simulation example (see Figure\,\ref{fig:metrics_2}). 
        
        The \ac{rmse} between the true measurements and the estimated output variables is given in the right plot of Figure\,\ref{fig:metrics_2}. 
        Although we discard the initial $10,000$ samples as burn-in, we see a significantly faster convergence after at most $1,000$ iterations. 
        
		\clearpage
		\section{Additional Simulation Results}\label{sup:wo_symmetry_sontraint}
\begin{figure}[h]
			\centering
			\newcolumntype{C}[1]{>{\centering\arraybackslash}p{#1}}
			
			
			{\fontsize{7pt}{7pt}\selectfont
				\begin{tabular}{p{0.225\textwidth} | C{0.225\textwidth}  C{0.225\textwidth} C{0.225\textwidth}}
					\toprule[1pt]
					\myfont{Hamiltonian model} &  \myfont{Reduced-rank \acs{gp}}   & \myfont{Reduced-rank \acs{gp}} & \myfont{Reduced-rank \acs{gp}} \\
					\midrule
					Symmetry constraints & None & Symmetry & None \\
					Training data  & \textcolor[RGB]{120,0,0}{input-state data} & \textcolor[RGB]{0,120,0}{input-output data} & \textcolor[RGB]{0,120,0}{input-output data} \\
					Flow magnitude \ac{rmse}  & $3.09\,\mathrm{J}$ & $3.27\,\mathrm{J}$ & $2.48\,\mathrm{J}$ \\
					Flow angle error \ac{rmse} & $0.08\,\mathrm{rad}$ & $0.12\,\mathrm{rad}$ & $1.47\,\mathrm{rad}$  \\ 
					\midrule
					\resizebox{\figwidthnew}{!}{} &
					\includegraphics[width=\figwidthnew]{img/Fig1_vp_is_bayes.pdf} &
					\includegraphics[width=\figwidthnew]{img/Fig1_vp_io_bayes_odd_nd_10000.pdf} &
					\includegraphics[width=\figwidthnew]{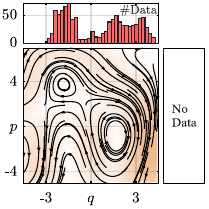} \\
					\bottomrule[1pt]
			\end{tabular}}
			\caption{Flow maps following from the true and learned Hamiltonians of the non-harmonic oscillator system.}
			\label{fig:Hamiltonians_supplementaries}
		\end{figure}

		Due to identifiability issues in the input-output setting, we impose a symmetry constraint on the Hamiltonian \ac{gp} model \eqref{eq:rr_gp_pred}. 
		To this end, an advantage of the employed reduced-rank \ac{gp} is that prior knowledge about the target function, such as symmetry, can be easily encoded in the basis function expansion \citep{Volkmann.2025}. 
		To impose the symmetry constraint, we follow the lines of \cite{Berntorp.2022} and index basis functions in \eqref{eq:basis_functions} by selecting only odd indices $j_{k,i} = 1, 3, \hdots$ when constructing the reduced-rank \ac{gp}. 
		This ensures that the approximated Hamiltonian satisfies $\widehat{H}(\bi{x}) = \widehat{H}(-\bi{x})$, which induces the anti-symmetry \mbox{$\nablabx \widehat{H}(\bi{x}) = - \nablabx \widehat{H}(-\bi{x})$} for its gradient. 
		In Figure\,\ref{fig:Hamiltonians_supplementaries}, we compare the proposed method in three different settings, \ie learning with
		\begin{itemize}
			\item state measurements $\{\bi{y}^{(\mathrm{i/s})}_t\}_{t=0}^T$, without symmetry constraint (left column),
			\item output measurements $\{y^{(\mathrm{i/o})}_t\}_{t=0}^T$, with symmetry constraint (middle column), and
			\item output measurements $\{y^{(\mathrm{i/o})}_t\}_{t=0}^T$, without symmetry constraint (right column).
		\end{itemize}
		
		All approaches correctly learn the equilibrium positions along the $q$-dimension. In the input-output setting, \ie without measurements of the momentum $p$ available, learning along the $p$-dimension can result in ambiguous solutions, which we attribute to identifiability problems. 
		In the present nonlinear simulation example, the symmetry constraint can facilitate learning from input-output data.
		
	\end{document}